\documentclass{article}

\PassOptionsToPackage{numbers, compress}{natbib}

\usepackage[final]{neurips_data_2022}

\usepackage[utf8]{inputenc} 
\usepackage[T1]{fontenc}    
\usepackage{hyperref}       
\usepackage{url}            
\usepackage{booktabs}       
\usepackage{amsfonts}       
\usepackage{nicefrac}       
\usepackage{microtype}      
\usepackage{xcolor}         
\usepackage{graphicx}
\usepackage{makecell}
\usepackage{multirow}
\usepackage{footmisc}

\usepackage{amssymb}
\usepackage{amsmath}
\usepackage{mathtools}
\usepackage{amsthm}
\usepackage{subcaption}

\newcommand{\ADD}[1]{{#1}}

\newcommand{\etal}{\textit{et al}. }

\title{Understanding Aesthetics with Language: A Photo Critique Dataset for Aesthetic Assessment}

\author{%
  Daniel Vera Nieto
  \\
  Media Technology Center (MTC) \\
  ETH Zurich \\
  Zurich, CH \\
  \texttt{daniel.veranieto@inf.ethz.ch} \\
  \And
   Luigi Celona \\
   DISCo \\
   University of Milano-Bicocca \\
   Milano, IT \\
   \texttt{luigi.celona@unimib.it} \\
   \AND
   Clara Fernandez-Labrador \\
   Media Technology Center (MTC) \\
   ETH Zurich \\
   Zurich, CH \\
  \texttt{clabrador@inf.ethz.ch}
}

\begin{document}

\maketitle

\begin{abstract}
Computational inference of aesthetics is an ill-defined task due to its subjective nature. Many datasets have been proposed to tackle the problem by providing pairs of images and aesthetic scores based on human ratings. However, humans are better at expressing their opinion, taste, and emotions by means of language rather than summarizing them in a single number. In fact, photo critiques provide much richer information as they reveal how and why users rate the aesthetics of visual stimuli. In this regard, we propose the Reddit Photo Critique Dataset (RPCD), which contains tuples of image and photo critiques. RPCD consists of 74K images and 220K comments and is collected from a Reddit community used by hobbyists and professional photographers to improve their photography skills by leveraging constructive community feedback. The proposed dataset differs from previous aesthetics datasets mainly in three aspects, namely (i) the large scale of the dataset and the extension of the comments criticizing different aspects of the image, (ii) it contains mostly UltraHD images, and (iii) it can easily be extended to new data as it is collected through an automatic pipeline. To the best of our knowledge, in this work, we propose the first attempt to estimate the aesthetic quality of visual stimuli from the critiques. To this end, we exploit the polarity of the sentiment of criticism as an indicator of aesthetic judgment. We demonstrate how sentiment polarity correlates positively with the aesthetic judgment available for two aesthetic assessment benchmarks. Finally, we experiment with several models by using the sentiment scores as a target for ranking images. Dataset and baselines are available\footnote{\url{https://github.com/mediatechnologycenter/aestheval}}.

\end{abstract}

\section{Introduction}
\label{sec:Intro}
Automated Image Aesthetic Assessment (IAA) is a widely discussed topic in the computer vision community and is receiving an increasing attention due to the explosive growth of digital photography. The literature reports mainly on predicting aesthetic preference in close agreement with human judgement. In particular, most studies deal with IAA in terms of high vs. low aesthetic quality \cite{cuhk-pq}, regression of the aesthetic score \cite{hosu2019effective}, and prediction of the distribution of the aesthetic ratings \cite{talebi2018nima}. Various datasets were collected to contribute developing and evaluating the previous studies. These datasets consist of images annotated with aesthetic scores. However, summarizing the aesthetic judgment in a single value limits the representation of visual aesthetics. First, aesthetic scores are highly dependent on the voting procedure (i.e., voting scale, number of stimuli, questions and adjectives in the voting scale). Second, it has been shown that they might provide a variable or even negative impact on the prediction of human behavior and thus on the success of social content \cite{rocklage2021mass}. Third, aesthetic scores do not provide any interpretability of why an image is aesthetically pleasing or not. Thus, it makes sense to annotate images with richer high-level aesthetic attributes \cite{kong2016photo} or aesthetic criticism captions. Many image-sharing sites, e.g., Flickr, Photo.net, and Instagram, support user comments on images, allowing rating explanations. User comments usually introduce rationale about how and why users evaluate the aesthetics of an image. Comments such as ``good composition'', ``vivid colors'', or ``a fine pose'' are more informative than ratings for expressing pleasing photographic aspects. Similarly, comments such as ``too dark'' and ``blurry'' explain why low ratings occur.

On the basis of previous considerations, our first contribution is the Reddit Photo Critique Dataset \textbf{RPCD}, a collection of high resolution images associated with photo critiques (i.e., 74K images and 220K comments). The dataset has been obtained from a Reddit community\footnote{\url{www.reddit.com}} of photography amateurs whose purpose is to provide feedback to help amateurs and professional photographers improve. Figure \ref{fig:rpcd_samples_with_sentscore} shows some samples from our RPCD. The dataset presented in this work differs in many ways from existing photo critic datasets. First, the images are mostly FullHD as they were captured with recent photo sensors and imaging systems. Secondly, the proposed dataset is among the largest in terms of the number of images-comments. Third, the comments of our dataset are on average longer and more informative (basing on the score proposed in \cite{ghosal2019aesthetic}) than those of the previous datasets. 

\ADD{In the literature, IAA models are trained on datasets in which each image is associated with a rating. However, the problem of how to obtain a rating for IAA without requiring human intervention given a dataset with image-comment pairs is not addressed.} The degree of emotion and valence of critique comments is an excellent indicator of the success of contents on social media \cite{rocklage2021mass}. Therefore, together with the dataset, we present a new solution to rank images by exploiting the polarity of criticism as an indicator of aesthetic judgments. To the best of our knowledge, this is the first attempt to leverage image critiques to define a score for IAA.

Finally, we design a framework to evaluate different methods on the proposed dataset and other aesthetic critique datasets in the literature on the image aesthetic assessment and aesthetic image captioning tasks.

We find that: (i) the aesthetic scores and the proposed sentiment scores are positively correlated on two photo critique datasets annotated with both comments and scores; (ii) Vision Transformer (ViT) surpasses state-of-the-art methods for image aesthetic assessment; (iii) learning aesthetics-aware features produces a significant increase in performance over using semantic features. This behavior also occurs for models with the same architecture but trained for different purposes.
\section{Related Work}
In this section we briefly analyze the main datasets and methods for the two tasks of image aesthetic assessment and aesthetic critique captioning. We refer the reader to \cite{zhang2021comprehensive} for a comprehensive review on computational image aesthetics.
\paragraph{Image Aesthetic Assessment.}
For the design and evaluation of Image Aesthetic Assessment (IAA) methods, the construction of the aesthetic image evaluation benchmark dataset has become the fundamental prerequisite for the research. Many datasets were collected in which subjective aesthetic quality scores were acquired for each image. The acquisition of subjective scores can be realized through manually scoring experiments in the lab~\cite{cuhk-pq}, online scoring on image sharing website~\cite{kong2016photo,murray2012ava}, and crowdsourcing evaluation~\cite{schifanella2015image}.
Methods that exploit the previous datasets for aesthetic assessment can be divided into model-based \cite{datta2006studying,marchesotti2011assessing,zhang2014fusion} and data-driven \cite{celona2021composition,hosu2019effective,lu2014rapid,talebi2018nima}. While model-based methods rely on hand-crafted features to model aspects such as the Rule of Thirds, depth of field, colour harmony, etc., the data-driven methods usually train CNNs on large-scale datasets to predict an overall aesthetic rating.
\paragraph{Aesthetic Critique Captioning.}
The first work on Aesthetic Critique Captioning, also known as Aesthetic Image Captioning (AIC), presents the so called Photo Critique Captioning Dataset (PCCD) based on a professional photo critique website\footnote{\url{https://gurushots.com/}} and a method for predicting aspect-centric captions \cite{chang2017aesthetic}.
The other AIC datasets in the literature are obtained by crawling images together with their comments from an on-line community of photography amateurs\footnote{\url{https://www.dpchallenge.com/}}. AVA-Comments \cite{zhou2016joint} extends AVA to include all user comments for images, while AVA-Captions \cite{ghosal2019aesthetic} filters original AVA photo comments to keep only the most useful. 
Finally, DPC-Captions \cite{jin2019aesthetic} contains 154,384 images and 2,427,483 comments. Each comment is automatically annotated with one of the 5 aesthetic attributes of the PCCD through aesthetic knowledge transfer. Few AIC methods are present in the literature for predicting aesthetic comments \cite{ghosal2019aesthetic,yeo2021generating}, aspect-centric aesthetic captions \cite{chang2017aesthetic}, or simultaneously the aesthetic rate and an aesthetic caption \cite{wang2019neural}.

\section{Background and Theory}

In this section we provide a formal definition for the classical image aesthetic assessment problem and describe our novel formulation of the problem.

\paragraph{Notations.}
We represent sets and matrices with special Latin characters (e.g., $\mathcal{M}$) or bold Latin characters (e.g., $\mathsf{M}$). Lower or uppercase normal fonts, e.g., $K$ denote scalars. Lowercase bold Latin letters represent vectors as in $\mathsf{v}$. We use lowercase Latin letters to represent indices (e.g., $i$). 

\subsection{Image Aesthetic Assessment}
Image Aesthetic Assessment (IAA) methods aim at computationally judging the aesthetic value of images based on human ratings and photographic principles. These methods map an input image $I_i \in \mathbb{R}^{H\times W\times3}$ to an aesthetic score $s_i$ and can be divided into binary classification methods which predict a single binary score $s \in \{0,1\}$, and regression methods which predict a single real score $s \in \mathbb{R}$ or a probability distribution of scores $p(s)$. Classification methods are used to distinguish ``good'' from ``bad'' images whereas regression methods are preferred to rank collections of images. These methods typically rely on public datasets \cite{kong2016photo,cuhk-pq,murray2012ava} that contain $N$ pairs of images and aesthetic scores such that $\mathcal{D}=\{(I_1, s_1), \hdots, (I_N, s_N)\}$, where the ground truth score per image is computed as the average rating given by $K$ human raters:

\begin{equation}
\label{eqn:avgscore}
    s_i = \frac{1}{K} \sum_{k=0}^K s_k,
\end{equation}

The scores are further thresholded by the mid-point of the rating scale for the classification task. However, asking people to evaluate the aesthetic value of an image with a single global score is very challenging and can be extremely biased by the content of the images. Additionally, these scores alone do not provide any explicit information about the reasons behind the voting.


\subsection{Aesthetic Critiques}
Recent datasets \cite{chang2017aesthetic, ghosal2019aesthetic,jin2019aesthetic} extend the IAA problem including captions related to photo aesthetics and/or photography skills. These datasets contain $N$ images each described by $K$ aesthetic critiques $\mathsf{c}$ such that $\mathcal{D}=\{(I_1, \mathsf{c}_1^1, \hdots, \mathsf{c}_1^K) \hdots, (I_N, \mathsf{c}_N^1, \hdots, \mathsf{c}_N^K)\}$. Common critiqued aesthetic aspects are composition, subject of photo, use of camera or color. In this context, novel algorithms have been developed to generate aesthetic-oriented critiques for images. Therefore, these methods map an input image $I_i \in \mathbb{R}^{H\times W\times3}$ to an aesthetic critique $\mathsf{c}_k$. While photo critiques give explicit feedback about why images are aesthetically pleasing or not, it has not been explored yet how to exploit such critiques for classification or image ranking, which is the ultimate goal of IAA. Furthermore, critique generative models present an additional challenge with respect to conventional captioning models due to the subjective nature of problem at hand. For this reason, many generated critiques tend to express critic's preference (e.g., ``I like the colors'' and ``nice photo'') rather than providing a detailed opinion or critique of the image aesthetics.
\subsection{Leveraging Aesthetic Critiques for Image Aesthetic Assessment} \label{sec:sentimentheory}
Ranking images from aesthetic critiques comes as a natural extension of the problem. In this paper, we are interested in leveraging the interpretability given by image captions to automatically discover the aesthetic score that truly defines the images beauty. Given an input image $I_i \in \mathbb{R}^{H\times W\times3}$ and $K$ aesthetic critiques associated to the image, we propose to use sentiment polarity analysis on each critique $\mathsf{c}_k$ to define the aesthetic score $s_i$ of the image. Sentiment polarity for a comment defines the orientation of the expressed sentiment, i.e., it determines if the text expresses the negative, neutral or positive sentiment of the user about the entity in consideration. A sentiment polarity model maps a given critique $\mathsf{c}_k$ to a vector $\mathsf{p}\in\mathbb{R}^{3}$ which can be interpreted as the probabilities of the given critique to express a negative, neutral or positive feeling with respect to the aesthetic value of the image. We define the sentiment score $s_k$ of a critique as follows:

\begin{equation}
    s_k = \frac{\sum_{l=0}^2 p_l l}{2},
\end{equation} 

Where $l$ is the label for negative, neutral or positive sentiment respectively and $p_l$ the probability associated to the label. The sentiment scores for all the critiques of an image are then averaged to obtain an overall sentiment score $s_i$ as in Eq. \ref{eqn:avgscore}. The proposed dataset can then be defined such that $\mathcal{D}=\{(I_1, \mathsf{c}_1^1, \hdots, \mathsf{c}_1^K, s_1) \hdots, (I_N, \mathsf{c}_N^1, \hdots, \mathsf{c}_N^K, s_N)\}$.

To the best of our knowledge, this is the first attempt to estimate the aesthetic quality of visuals directly from critiques rather than from human ratings. \ADD{Our proposal is significant for several reasons. Critiques are an important indicator of human judgment, generally more valuable than simple ratings as they provide an explanation of why a visual is aesthetically pleasing or not \cite{rocklage2021mass}. However, critiques are unstructured data that do not directly indicate the level of aesthetic appreciation. Therefore, our proposed score is a way to obtain a compact and quantifiable representation of the level of appreciation of an image inferred from the critiques. Second, thanks to our proposal, two related aesthetic tasks are linked. Indeed, the datasets created for Aesthetic Image Captioning (AIC) can be applied to the design of models for both AIC and Image Aesthetic Assessment (IAA). The integration of the two tasks is useful because the critiques guarantee the explainability of a score; on the other hand, the ratings might allow the prediction of valence-sensitive critiques. Finally, our proposal consists of a weakly-supervised labeling approach which has the advantage of requiring human intervention solely to provide comments on the image. Existing datasets such as PCCD and AVA demanded intensive human effort to provide ratings and comments.}
\section{RPCD: Reddit Photo Critique Dataset}
The Reddit Photo Critique Dataset (RPCD) is a collection of high-resolution images associated with photo critiques obtained by the Reddit communities. We first give a description of the dataset collection and statistics in Section \ref{sec:dataset-info}. Section \ref{sec:dataset-sentiment-scoring} details how we automatically rank the images following the criticism-based approach described in Section \ref{sec:sentimentheory}. Finally, in Section \ref{sec:analysis}, we thoroughly analyze the images and comments present in our dataset.
\subsection{Dataset Collection and Statistics}
\label{sec:dataset-info}
\paragraph{Collection Modality.}
For the collection of the RPCD dataset, we identified Reddit communities used by amateur and professional photographers to upload their images or to discuss about photography. In particular, the following six communities (known as subreddits) were identified: \texttt{/r/AskPhotography}, \texttt{/r/photocritique}, \texttt{/r/photographs}, \texttt{/r/portraits}, \texttt {/r/postprocessing}, \texttt{/r/shittyHDR}. After a careful review of the different subreddits, we selected the \texttt{/r/photocritique} subreddit with a total of 168,222 posts and 731,772 comments. The decision was made based on the rules of the community\footnote{\url{https://www.reddit.com/r/photocritique/}} which makes its content specially suitable for the task at hand. Namely, it mostly contains posts with amateur and professional images that get feedback from other photographers and hobbyists. We downloaded all the posts and comments from the selected subreddit between May 2009 and February 2022 using the Python Reddit API Wrapper (PRAW\footnote{\url{https://github.com/praw-dev/praw} (Accessed on 06/05/2022).}) and the Pushshift platform~\cite{baumgartner2020pushshift}. Nevertheless, we still note that the other subreddits may hold relevant information that can also be exploited in further works. See Appendix \ref{app:subanaly} for more details regarding the number of posts/comments per year in the aforementioned subreddits.

\paragraph{Automatic Filtering.} The selected posts are then filtered by using an automated pipeline designed to be reused over time or for other communities. It consists of the following steps. First, each post consists of an image along with a description provided by the photographer usually explaining the aesthetic intent of the photograph and the technical details of the camera used. Additionally, each post has comments from other users structured as layered conversations. As required by the subreddit rules, the first level comments must be a critique to the image in the post. Therefore, we keep the top level comments since they are actual critiques and they are not a follow up comment or part of the body of a conversation. The description and the comments under the first level are discarded, thus reducing the number of comments from 731,772 to 284,426. Secondly, we remove the posts with no comments or whose image is no longer available. As a result, the number of posts is reduced to 103,190. Finally, filtering posts with corrupted or placeholder images leads to the final dataset consisting of 73,965 posts, each of them consisting of an image and an average of 3 critiques to that image.

\paragraph{Statistics.} Our RPCD dataset consists of 73,965 images with a resolution of $2993\times 2716$ pixels on average. A total of 219,790 photo critique comments is available, with an average of 49.1 words per comment, a standard deviation of 55.5 and a maximum of 1286 words. Each image has an average of 2.9 comments associated with it, with a standard deviation of 3.7. The general information of our dataset and a comparison with related datasets is presented in Table \ref{tab:general}. Several considerations can be made. First, our RPCD dataset is the first large-scale photo critique dataset, with a $\thicksim\!17$ times and $\thicksim\!7$ times increase in the number of images and comments, respectively, compared to the previously available photo critique dataset, PCCD \cite{jin2019aesthetic}. Secondly, it has a slightly higher average length of comments with respect to PCCD and about 3 times that of AVA-Comments \cite{zhou2016joint} (hereafter simply referred to as AVA). This increase in the amount of information opens the door to the use of large language models to exploit the unstructured information available in form of text. Third, our RPCD dataset consists of images with a much higher resolution than previous datasets (especially those obtained from \url{DPChallenge.com}). See Appendix \ref{app:imageres} for a detailed comparison. This may be due to the difference in the time periods of collection for the different datasets. For example, AVA dataset has images posted only until 2011, when the technical performance and availability of cameras were inferior to nowadays. Consequently, the aesthetic quality is very likely to depend on the perceived technical quality \cite{kang2020eva}.
\begin{table}
    \centering
    \caption{Comparison of the properties in different benchmark datasets on image aesthetic captioning.}
    \label{tab:general}
    \resizebox{\textwidth}{!}{\begin{tabular}{lcccc}
        \toprule
        Dataset & \makecell{AVA-Comments\\ \cite{zhou2016joint}} & \makecell{DPC-Captions$^{**}$\\ \cite{jin2019aesthetic}} & \makecell{PCCD\\ \cite{chang2017aesthetic}} & \makecell{RPCD \\(Our)} \\ \midrule
        Images & 253,961 & 117,132 & 4,235 & 73,965 \\
        Avg image resolution & 607$\times$537 & 606$\times$534 & 1414$\times$1202 & 2993$\times$2716 \\
        Attributes & -- & 5 & 7 & 7$^*$ \\
        Comments & 3,601,761 & 208,926 & 29,645 & 219,790 \\
        Comments per image & 14.1 & 1.8 & 6.6 & 2.9\\
        Avg words per comment & 14.6 & 24.5 & 41.1 & 49.1 \\
        Max words per comment & 2146 & 549 & 780 & 1286 \\
        Content category & 66 & 66 & 27 & 6$^*$ \\
        Rating scale & 1-10 & 1-10 & 1-10 & 0-1$^*$ \\
        Avg raters per image & 6 & 15 & 7 & -- \\
        \bottomrule
    \end{tabular}}
    \begin{flushleft}{\footnotesize$^*$The aspect is obtained through machined-based annotation. See Appendix~\ref{appx:content}.\\
    $^{**}$The figures reported on this table are produced using the code made available by the authors of the dataset and differ from those stated in the original paper.}\end{flushleft}
\end{table}
\subsection{Sentiment polarity prediction}
\label{sec:dataset-sentiment-scoring}
We propose to use sentiment polarity analysis on the aesthetic critiques to define the aesthetic scores, a.k.a sentiment scores, of the images as detailed in Section \ref{sec:sentimentheory}. \ADD{Sentiment analysis methods can be categorized into lexicon-based methods \cite{huq2017sentiment,salas2017sentiment}, machine learning methods \cite{gowda2022sentiment,zhang2016comparison}, and hybrid methods \cite{pandey2017twitter}. Recently, deep learning methods have enabled the design of sentiment analysis models that have achieved impressive performance over the previous methods \cite{bingyu2022document,DBLP:journals/corr/abs-1907-11692,punith2021sentiment,rosenthal-etal-2017-semeval}.}

\begin{figure}
    \centering
    \includegraphics[width=\linewidth]{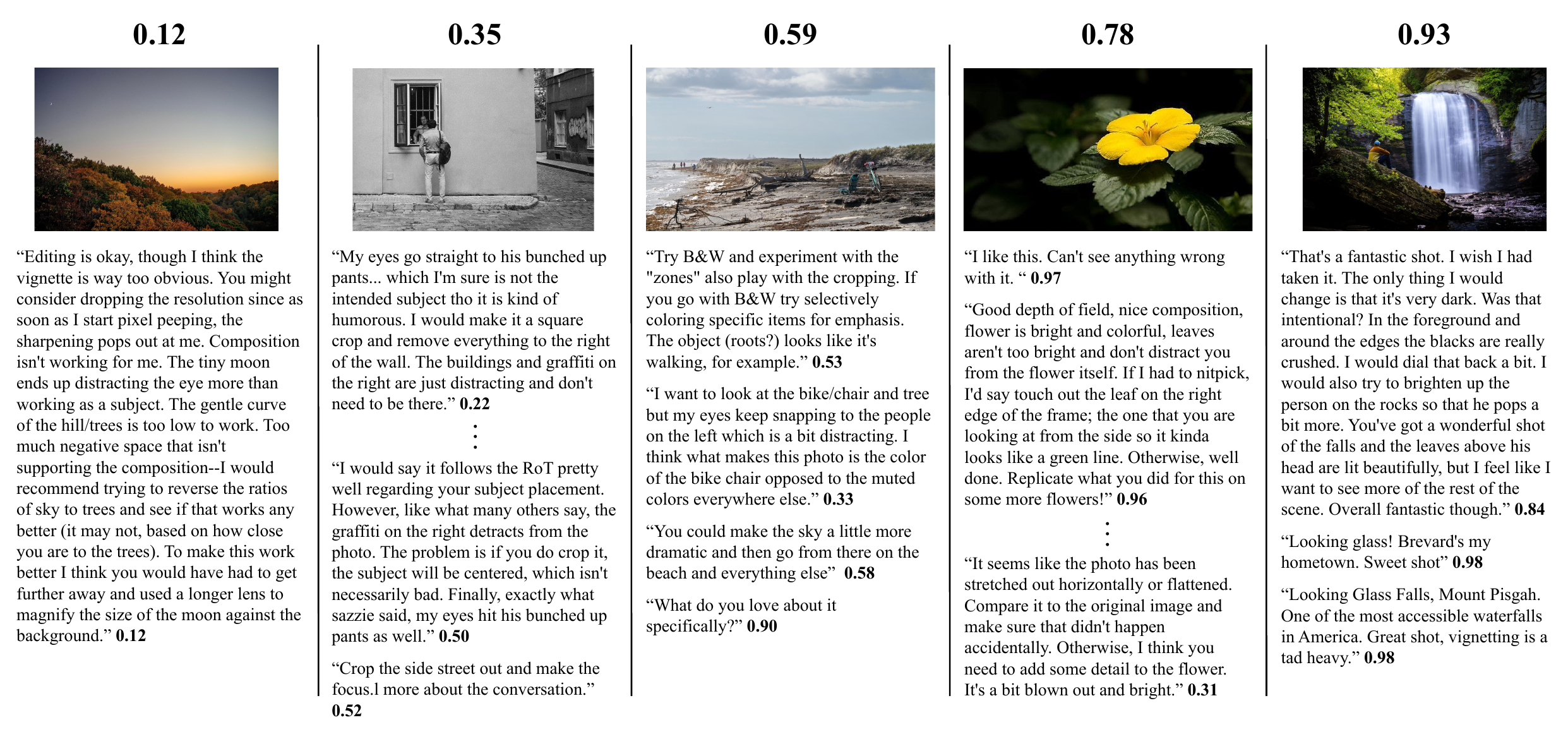}
    \caption{RPCD samples annotated with the proposed sentiment score. Sentiment scores are also reported for each comment.}
    \label{fig:rpcd_samples_with_sentscore}
\end{figure}

\begin{figure}
    \centering
    \includegraphics[width=\textwidth]{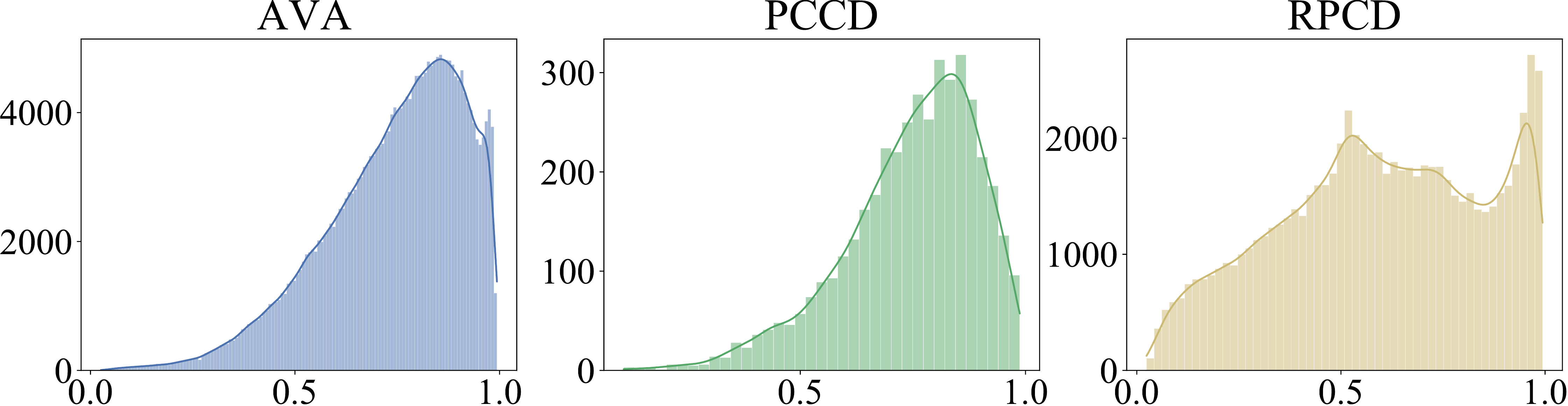}
    \caption{Sentiment score distribution on AVA, PCCD and our RPCD dataset.}
    \label{fig:reddit_score_dist}
\end{figure}

\ADD{In this work, we use TwitterRoBERTa~\cite{barbieri-etal-2020-tweeteval}, a deep learning based method inspired by RoBERTa~\cite{DBLP:journals/corr/abs-1907-11692}, to extract the sentiment polarity on aesthetic critiques. TwitterRoBERTa achieved the best trade-off between performance and model complexity among all the models that participated in the Sentiment Analysis in the Twitter challenge~\cite{rosenthal-etal-2017-semeval}. Although the model is trained in a different domain (Twitter), we assume that the domain is similar enough (social media) to use this model. Future work could explore the use of models tailored for the Reddit sub-domain. Additionally, Transformer models fine-tuned for sentiment analysis are known to have their own set of bias~\footnote{\url{https://huggingface.co/distilbert-base-uncased-finetuned-sst-2-english\#risks-limitations-and-biases}}. Hence, a deeper analysis of the bias introduced by the selected model, which was not present in the original work~\cite{barbieri-etal-2020-tweeteval}, would be beneficial.} We exploit the implementation of TwitterRoBERTa finetuned for the sentiment prediction task vailable in the HuggingFace transformers library~\cite{https://doi.org/10.48550/arxiv.1910.03771}. Figure \ref{fig:rpcd_samples_with_sentscore} reports some samples of our dataset annotated with sentiment scores. Individual scores per critique are also included. It can be seen that most of the comments are focused on compositional and stylistic aspects of the photo.
We estimate the sentiment score of the comments of AVA and PCCD for comparison. Figure \ref{fig:reddit_score_dist} shows the distribution of sentiment scores for the samples of AVA, PCCD and our RPCD. We observe that the vast majority of the AVA and PCCD dataset samples are characterized by high sentiment scores, which produce left-skewed score distributions. On the other hand, the samples of the proposed RPCD cover almost the entire range of values with two peaks close to the values 1 and 0.5. This difference between ours and the other datasets indicates that RPCD have a richer representation of the whole aesthetic taste spectrum, providing information about why an image have a specific score for high and low sentiment scores. This dissimilarity can be explained by the nature of the different sources of each dataset. While \texttt{DPChallenge} is a community where users score each image, they are not encouraged to critique them as in the \texttt{r/photocritique} subreddit. Consequently, we hypothesize, this produces that only those users with a praise would leave a comment. The fact that there are many more users giving a score than commenting supports this possible explanation. The case of the PCCD dataset is more difficult to analyze since the source website\footnote{\url{https://gurushots.com/}} does not provide the critiques feature anymore, but the fact that the dataset is heavily imbalanced may explain why the sentiment score is also imbalanced.

\begin{figure}
    \centering
    \begin{tabular}{cc}
        \includegraphics[width=.45\textwidth]{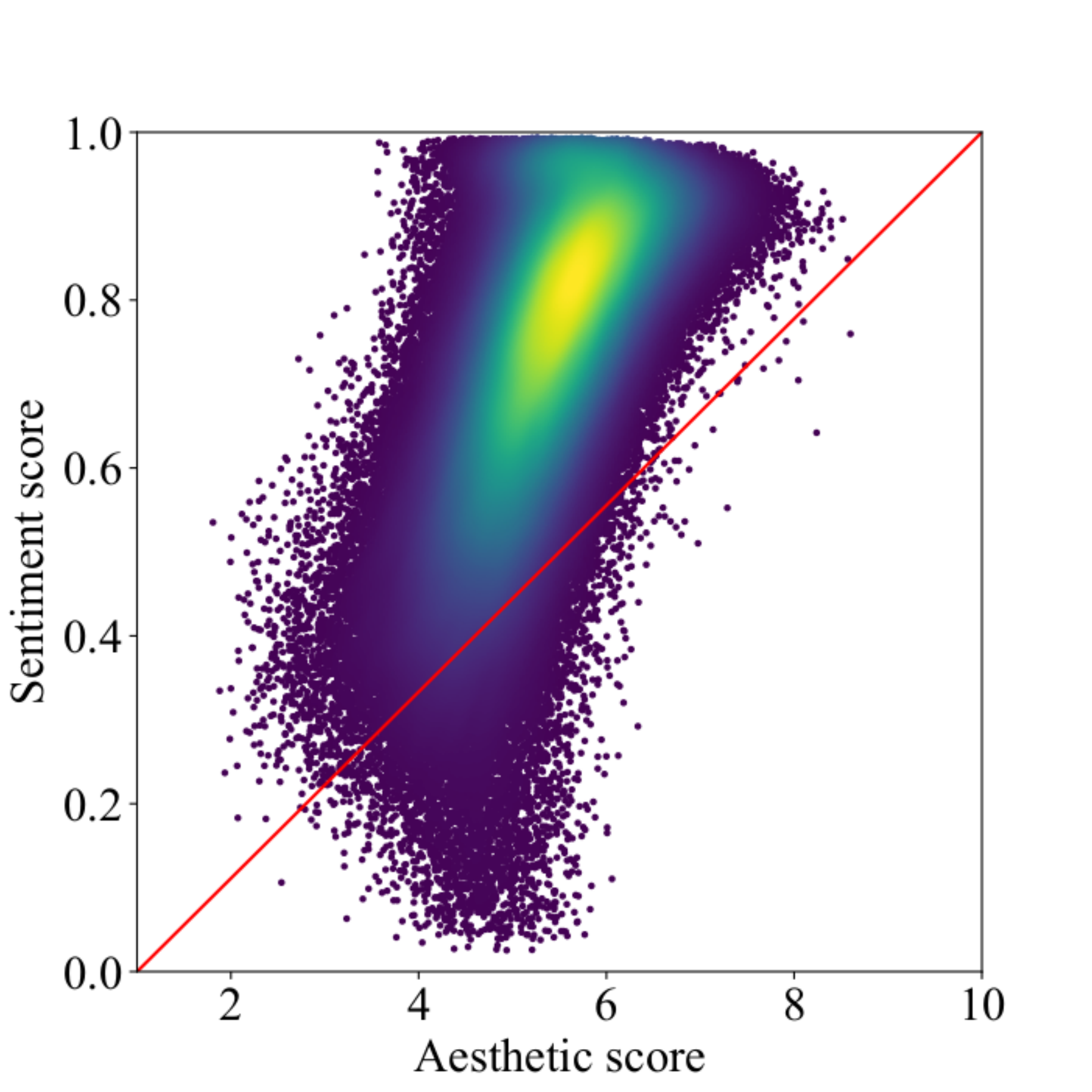} &
        \includegraphics[width=.45\textwidth]{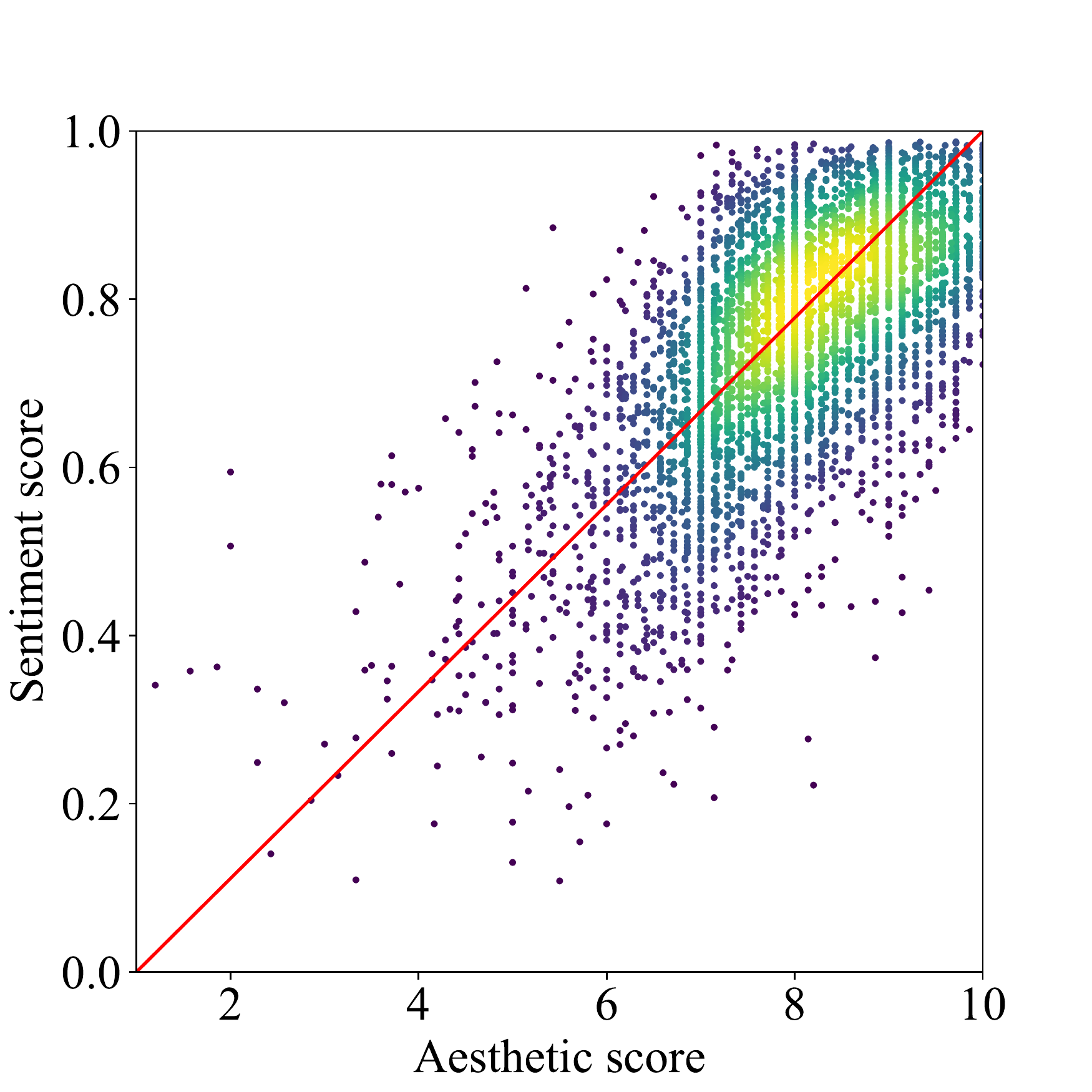} \\
        (a) AVA & (b) PCCD 
    \end{tabular}
    \caption{Annotated aesthetic score \textit{vs.} Sentiment polarity score for (a) AVA and (b) PCCD samples.}
    \label{fig:correlation}
\end{figure}

We also estimate how the sentiment score correlates with the annotated human aesthetic judgment for the AVA and PCCD images. In particular, we measure Spearman's Rank Correlation Coefficient (SRCC) and Pearson's Linear Correlation Coefficient (PLCC). On the AVA dataset, the SRCC is equal to 0.6418 while PLCC corresponds to 0.6424. For PCCD, SRCC is 0.6066 and PLCC 0.6499. The positive correlation on both datasets indicates the effectiveness of the proposed score and, therefore, that it represents a trustworthy approximation of the aesthetic score. Figure \ref{fig:correlation} shows the scatter plots relating the aesthetic score and sentiment score for the two considered datasets. It can be seen that most of the AVA aesthetic scores were originally annotated around the average value of the rating scale, i.e. 5. In fact, it is worth noting that AVA sentiment scores span the whole rating scale for aesthetic scores equal to or close to 5. We deepen the latter case in Appendix \ref{app:subanaly}. PCCD original scores, on the other hand, are very positive with a high concentration of samples for values between 7 and 9. Generally, our sentiment scores take on less biased values than previous aesthetic scores.
\subsection{Content Analysis}
\label{sec:analysis}
In this section, we present an in-depth analysis of the images and comments in our dataset. This analysis is conducted by training different models to annotate aspects related to the semantics and composition of images and to estimate the usefulness and topics of the comments.

\begin{figure}
    \centering
    \includegraphics[width=\textwidth]{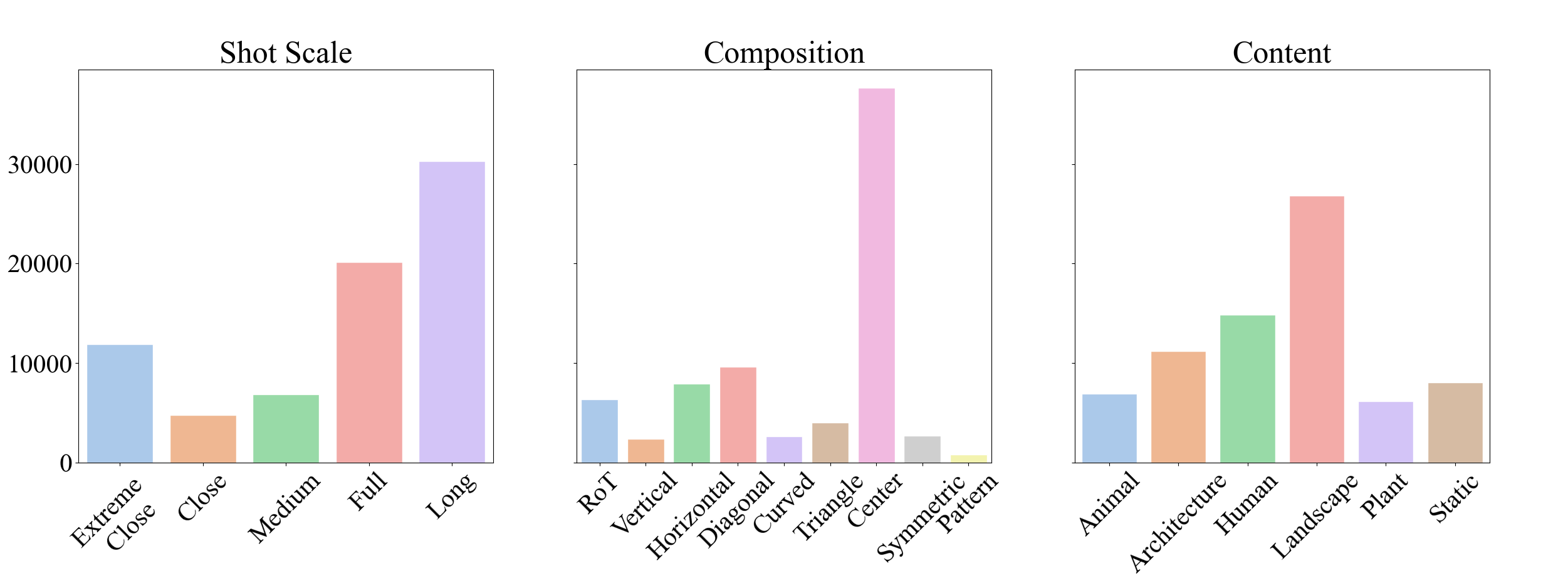} 
    \caption{RPCD analysis. Left: Shot scale. Center: Image Composition, Right: Image Content.}
    \label{fig:annotations}
\end{figure}

\paragraph{Image Analysis.} For semantic content analysis, we group images into six categories, i.e., Animal, Architecture, Human, Landscape, Plant, and Static/Others. The semantic categories above are the same as CUHK-PQ, excluding the Night category. The latter is misleading as it represents the time of the shot and not a semantic category. Figure \ref{fig:annotations} shows the distribution of images per semantic category. As it is possible to see, most RPCD images contain landscapes. The second largest content category (approximately 15K images) includes human beings. The semantic class with the lowest number of instances is Plant. We also report results for shot scale classification, which determines the portion occupied by the main subject with respect to the frame. We distinguish five shot scale types defined in the training dataset MovieNet \cite{huang2020movienet}: Extreme close-up, Close-up, Medium, Full, Long. Figure \ref{fig:annotations} shows that about 30K images have been labeled as Long shot scale. This result is in line with the fact that most images are of the semantic Landscape type. Very few images have been classified as Close-up or Medium range content, meaning photographers have preferred to capture subjects from very close or far away. Finally, we inspect images from the point of view of photographic composition. Specifically, images are categorized with respect to the main composition rule among the following eight: Rule-of-Thirds, Vertical, Horizontal, Diagonal, Curved, Triangle, Center, Symmetric, Pattern. The above composition rules are defined in the KU-PCP dataset \cite{lee2018photographic}. Figure \ref{fig:annotations} presents the distribution of the images with respect to the composition rules. The images in our dataset are very bias on the Center category, indicating that in most images there is the main subject occupying the central region of the image. In Appendix \ref{appx:content} we detail how we build the models for running the previous analysis and present some sample images for each of the analyzed aspects.
\paragraph{Comment Analysis.} We analyze the topics of the corpus of comments in our dataset using BERTopic~\cite{grootendorst2022bertopic},  a topic modeling technique. The most common topics in the analyzed datasets regards semantic aspects such as face, tree, bird, flower, and stars. There are also topics related to technical aspects of photography, namely ISO, aperture, dynamic range, and HDR. We refer to Appendix~\ref{appx:topic} for a deeper analysis.
Additionally, we use the definition of informativeness score of a previous work~\cite{ghosal2019aesthetic} to estimate whether the comments of our dataset are meaningful and how do they compare to other datasets. In the Appendix~\ref{appx:info_score} we compare the results on our dataset with those of state-of-the-art datasets, finding that, on average, the proposed RPCD contains the most informative comments, with an informativeness score slightly higher than PCCD and more than double than AVA.
\section{Evaluation}
To illustrate the possible uses of the newly introduced dataset, we run several experiments around the image aesthetic assessment task, where our goal is to predict an aesthetic score given an image, as well as in the image captioning task, where our goal is to predict an aesthetic critique given an image. To this aim, we split the whole dataset into 70\% training samples, 10\% validation samples, and the remaining 20\% for testing.

\subsection{Image Aesthetic Assessment}
The main motivation to create this dataset is to perform Image Aesthetic Assessment with the interpretability of the aesthetic critiques. We use the scores computed using sentiment analysis and propose a method to predict such scores. We also run SOTA models on our dataset for comparison.



In Table \ref{tab:aesthetic-assessment-comparison} we report SRCC, LCC and Accuracy on our RPCD, PCCD \cite{chang2017aesthetic} and AVA~\cite{murray2012ava} datasets using the sentiment score, comparing the results of \texttt{NIMA} \cite{talebi2018nima} and other experiments we carried out. We additionally perform an extensive evaluation of the family of ViT models in Appendix~\ref{sec:implementation} to assess the suitability of such models for the aesthetic assessment task, evaluating its performance on AVA dataset. We highlight that ViT Large (we called \texttt{AestheticViT}) outperforms previous SOTA model~\cite{hosu2019effective} by a 4\% in the correlation metrics on AVA dataset using the original scores. In the \texttt{ViT + Linear probe} experiments, we also study to what extent the results obtained to predict the aesthetic and sentiment scores are due to the knowledge already present in the pretrained model. The accuracy is computed defining as high quality images those with an score above 5, and poor quality otherwise. The results in Table~\ref{tab:aesthetic-assessment-comparison} show that, although \texttt{ViT + Linear probe} and \texttt{AestheticViT} outperform a previous aesthetic model used as a baseline, it does not achieve a good enough performance in any of the benchmarks to predict the proposed sentiment score. Moreover, the training of the model on AVA deteriorate its performance. This results proofs how challenging the task is and may indicate that the main previous benchmark, AVA, may be biased towards the content of the images, reducing the importance of the actual aesthetics of the photo. We would like to support the reasoning of Hosu \etal \cite{hosu2019effective} regarding the suitability of correlation metrics rather than accuracy to evaluate this task. While correlation metrics are representative of the entire range of scores, image labels ('good' or 'bad') are defined arbitrarily, which becomes an issue when the label distribution is imbalanced as in the case of AVA and PCCD datasets for both the original and sentiment scores. 

\begin{table}[h]
  \caption{Sentiment score baseline on the three considered datasets.}
  \label{tab:aesthetic-assessment-comparison}
  \centering
  \resizebox{\linewidth}{!}{\begin{tabular}{l|ccc|ccc|ccc}
    \toprule
    \multirow{2}{*}{Method} & \multicolumn{3}{c|}{AVA} & \multicolumn{3}{c|}{PCCD} & \multicolumn{3}{c}{RPCD} \\
     & SRCC & LCC & Acc. (\%) & SRCC & LCC & Acc. (\%) & SRCC & LCC & Acc. (\%) \\
    \midrule
    NIMA \cite{talebi2018nima} & 0.253 & 0.259 & 90.20 & 0.066 & 0.070 & \textbf{93.87} & 0.120 & 0.116 & 63.25 \\
    ViT + Linear probe$^*$ & \textbf{0.570}  & \textbf{0.570} & 76.43 & 0.156& 0.165  & 93.04 & 0.172 & 0.173 & 64.58\\
    AestheticViT$^*$ & 0.544 & 0.550 & \textbf{90.54} & \textbf{0.228} & \textbf{0.262} & 93.86 & \textbf{0.250} & \textbf{0.253} & \textbf{65.27} \\
    \bottomrule
  \end{tabular}}
  \begin{flushleft}{\footnotesize$^*$Best performing models. See Appendix~\ref{sec:implementation}}\end{flushleft}
\end{table}
%
%
\subsection{Aesthetic Critiques}
We also evaluate our dataset on the task of Aesthetic Image Captioning (AIC), using a SOTA model \cite{li2022blip}. To evaluate the results, we follow the procedure of \cite{chang2017aesthetic} as our dataset also contains more than one reference caption that corresponds to a single image. Table \ref{tab:aesthetic-captioning-comparison} compares the obtained results with the previous work~\cite{chang2017aesthetic} we use as reference. We observe that the achievable performance is far lower than that obtained for the description of the image content (See COCO captions benchmark~\footnote{https://paperswithcode.com/sota/image-captioning-on-coco-captions}), and further work is necessary to produce meaningful aesthetic captions. More details on the aesthetic critique procedure can be found in Appendix~\ref{appx:aesth_critiques}.
\begin{table}[h]
    \centering
    \caption{Aesthetic image captioning using BLIP \cite{li2022blip} on PCCD and our RPCD.}
    \label{tab:aesthetic-captioning-comparison}
    \begin{tabular}{l|cccccccc}
    \toprule
         & Bleu1 & Bleu2 & Bleu3 & Bleu4 & METEOR & ROUGE & CIDEr & SPICE \\ \midrule
        PCCD & 0.165 & 0.065 & 0.028 & 0.011 & 0.063 & 0.137 & 0.049 & 0.048 \\
        RPCD & 0.211 & 0.088 & 0.038 & 0.017 & 0.077 & 0.157 & 0.048 & 0.040 \\
        \bottomrule
    \end{tabular}
\end{table}
%


\section{Discussion and Future Work}
\label{sec:discussion}
We presented the Reddit Photo Critique Dataset (RPCD) consisting of image and photo critiques tuples. This dataset is collected by crawling posts from a community where people are encouraged to criticize positive and negative image aesthetic aspects. Our dataset has approximately 18$\!\times$ the images and 7$\!\times$ the comments compared to the PCCD dataset. Compared to AVA, the best-known aesthetic assessment dataset, our RPCD has longer and more meaningful comments and higher resolution images. Together with the dataset, we have for the first time in the literature defined an approach to obtain the aesthetic ranking of images directly from the analysis of comments. The proposed approach is based on the sentiment analysis of the comments. The proposed score was shown to have a positive correlation with the aesthetic judgments of humans. Therefore, RPCD can be used both to predict aesthetic captions and to estimate an aesthetic score. We conducted several experiments for the image aesthetic assessment task in which we compared the results obtained from different methods on our dataset, AVA and PCCD. These experiments show that a ViT is able to obtain good performance on AVA while both PCCD and RPCD are more challenging. The use of content-aware (ViT + Linear Probe), instead of aesthetic-aware (Aesthetic ViT), features results in a significant drop in performance for those datasets that may be less content-biased. Experiments on aesthetic image captioning carried out on PCCD and RPCD datasets highlight that the achievable performance is far lower than that obtained for the description of the image content, and further work is necessary to produce meaningful aesthetic captions.

However, several limitations remain. First, the limited number of comments per image (i.e., 3 in average), although the comments are long and very informative, could make the evaluation biased by the few users and not sufficiently objective. Second, we encourage the Machine Learning community to work on alternative or complementary solutions to the proposed sentiment analysis as a proxy for aesthetics. Among other things, the fact that the sentiment score is automatically estimated could cause noisy annotations and the change in the data domain should be further studied. \ADD{This noisy annotations could be influenced by the potential bias present in the selected model}. Third, the aesthetic captioning task remains an open challenge. \ADD{Finally, the concept of aesthetics expressed in our dataset must be understood limitedly to the Western cultural and geographical context on the basis of the demographic statistics of the Reddit users (see Appendix~\ref{app:subanaly}). Additionally, other ethical considerations are discussed in Appendix~\ref{sec:ethics}}.

Despite the above limitations, we believe RPCD is an important contribution for the design of multi-modal and explainable aesthetic assessment models. As a future work, we would like to deepen the ranking method based on the analysis of comments in order to make it more reliable and diversified for the different aspects that characterize the aesthetics: style, color, composition, etc. In addition we will be able to interpret which are the aspects that are evaluated more positively or negatively by users. Finally, exploiting new sources of available data may provide further benefits while training larger models.

\section*{Acknowledgments}

This project is supported by Ringier, TX Group, NZZ, SRG, VSM, Viscom, and the ETH Zurich Foundation.

{
\small

\bibliographystyle{plainnat}
\bibliography{main}

\begin{thebibliography}{45}
\providecommand{\natexlab}[1]{#1}
\providecommand{\url}[1]{\texttt{#1}}
\expandafter\ifx\csname urlstyle\endcsname\relax
  \providecommand{\doi}[1]{doi: #1}\else
  \providecommand{\doi}{doi: \begingroup \urlstyle{rm}\Url}\fi

\bibitem[Barbieri et~al.(2020)Barbieri, Camacho-Collados, Espinosa~Anke, and
  Neves]{barbieri-etal-2020-tweeteval}
Francesco Barbieri, Jose Camacho-Collados, Luis Espinosa~Anke, and Leonardo
  Neves.
\newblock {T}weet{E}val: Unified benchmark and comparative evaluation for tweet
  classification.
\newblock In \emph{Findings of the Association for Computational Linguistics:
  EMNLP 2020}, pages 1644--1650. Association for Computational Linguistics,
  2020.

\bibitem[Baumgartner et~al.(2020)Baumgartner, Zannettou, Keegan, Squire, and
  Blackburn]{baumgartner2020pushshift}
Jason Baumgartner, Savvas Zannettou, Brian Keegan, Megan Squire, and Jeremy
  Blackburn.
\newblock The pushshift reddit dataset.
\newblock In \emph{International AAAI Conference on Web and Social Media},
  volume~14, pages 830--839, 2020.

\bibitem[Bingyu and Arefyev(2022)]{bingyu2022document}
Zhang Bingyu and Nikolay Arefyev.
\newblock The document vectors using cosine similarity revisited.
\newblock In \emph{Workshop on Insights from Negative Results in NLP}, pages
  129--133, 2022.

\bibitem[Celona et~al.(2022)Celona, Leonardi, Napoletano, and
  Rozza]{celona2021composition}
Luigi Celona, Marco Leonardi, Paolo Napoletano, and Alessandro Rozza.
\newblock Composition and style attributes guided image aesthetic assessment.
\newblock \emph{IEEE Transactions on Image Processing}, 31:\penalty0
  5009--5024, 2022.

\bibitem[Chang et~al.(2017)Chang, Lu, and Chen]{chang2017aesthetic}
Kuang-Yu Chang, Kung-Hung Lu, and Chu-Song Chen.
\newblock Aesthetic critiques generation for photos.
\newblock In \emph{International Conference on Computer Vision}, pages
  3514--3523. IEEE, 2017.

\bibitem[Chen et~al.(2020)Chen, Zhang, Zhou, Lei, Xu, Zheng, and
  Fan]{chen2020adaptive}
Qiuyu Chen, Wei Zhang, Ning Zhou, Peng Lei, Yi~Xu, Yu~Zheng, and Jianping Fan.
\newblock Adaptive fractional dilated convolution network for image aesthetics
  assessment.
\newblock In \emph{Computer Vision and Pattern Recognition}, pages
  14114--14123. IEEE/CVF, 2020.

\bibitem[Datta et~al.(2006)Datta, Joshi, Li, and Wang]{datta2006studying}
Ritendra Datta, Dhiraj Joshi, Jia Li, and James~Z Wang.
\newblock Studying aesthetics in photographic images using a computational
  approach.
\newblock In \emph{European Conference on Computer Vision}, pages 288--301.
  Springer, 2006.

\bibitem[Dosovitskiy et~al.(2021)Dosovitskiy, Beyer, Kolesnikov, Weissenborn,
  Zhai, Unterthiner, Dehghani, Minderer, Heigold, Gelly, Uszkoreit, and
  Houlsby]{dosovitskiy2020image}
Alexey Dosovitskiy, Lucas Beyer, Alexander Kolesnikov, Dirk Weissenborn,
  Xiaohua Zhai, Thomas Unterthiner, Mostafa Dehghani, Matthias Minderer, Georg
  Heigold, Sylvain Gelly, Jakob Uszkoreit, and Neil Houlsby.
\newblock An image is worth 16x16 words: Transformers for image recognition at
  scale.
\newblock In \emph{International Conference on Learning Representations}, 2021.

\bibitem[Gebru et~al.(2021)Gebru, Morgenstern, Vecchione, Vaughan, Wallach,
  Iii, and Crawford]{gebru2021datasheets}
Timnit Gebru, Jamie Morgenstern, Briana Vecchione, Jennifer~Wortman Vaughan,
  Hanna Wallach, Hal~Daum{\'e} Iii, and Kate Crawford.
\newblock Datasheets for datasets.
\newblock \emph{Communications of the ACM}, 64\penalty0 (12):\penalty0 86--92,
  2021.

\bibitem[Ghosal et~al.(2019)Ghosal, Rana, and Smolic]{ghosal2019aesthetic}
Koustav Ghosal, Aakanksha Rana, and Aljosa Smolic.
\newblock Aesthetic image captioning from weakly-labelled photographs.
\newblock In \emph{International Conference on Computer Vision Workshops},
  pages 0--0. IEEE/CVF, 2019.

\bibitem[Gowda et~al.(2022)Gowda, Archana, Shettigar, and
  Satyarthi]{gowda2022sentiment}
SR~Gowda, BR~Archana, Praajna Shettigar, and Kislay~Kumar Satyarthi.
\newblock Sentiment analysis of twitter data using na{\"\i}ve bayes classifier.
\newblock In \emph{ICDSMLA 2020}, pages 1227--1234. Springer, 2022.

\bibitem[Grootendorst(2022)]{grootendorst2022bertopic}
Maarten Grootendorst.
\newblock Bertopic: Neural topic modeling with a class-based tf-idf procedure.
\newblock \emph{arXiv preprint arXiv:2203.05794}, 2022.

\bibitem[Hong et~al.(2021)Hong, Du, Xian, Lu, Cao, and
  Zhong]{hong2021composing}
Chaoyi Hong, Shuaiyuan Du, Ke~Xian, Hao Lu, Zhiguo Cao, and Weicai Zhong.
\newblock Composing photos like a photographer.
\newblock In \emph{Conference on Computer Vision and Pattern Recognition},
  pages 7057--7066. IEEE/CVF, 2021.

\bibitem[Hosu et~al.(2019)Hosu, Goldlucke, and Saupe]{hosu2019effective}
Vlad Hosu, Bastian Goldlucke, and Dietmar Saupe.
\newblock Effective aesthetics prediction with multi-level spatially pooled
  features.
\newblock In \emph{Conference on Computer Vision and Pattern Recognition},
  pages 9375--9383. IEEE/CVF, 2019.

\bibitem[Huang et~al.(2020)Huang, Xiong, Rao, Wang, and Lin]{huang2020movienet}
Qingqiu Huang, Yu~Xiong, Anyi Rao, Jiaze Wang, and Dahua Lin.
\newblock Movienet: A holistic dataset for movie understanding.
\newblock In \emph{European Conference on Computer Vision}, pages 709--727.
  Springer, 2020.

\bibitem[Huq et~al.(2017)Huq, Ahmad, and Rahman]{huq2017sentiment}
Mohammad~Rezwanul Huq, Ali Ahmad, and Anika Rahman.
\newblock Sentiment analysis on twitter data using knn and svm.
\newblock \emph{International Journal of Advanced Computer Science and
  Applications}, 8\penalty0 (6), 2017.

\bibitem[Jin et~al.(2019)Jin, Wu, Zhao, Li, Zhang, Ge, Zou, Zhou, and
  Zhou]{jin2019aesthetic}
Xin Jin, Le~Wu, Geng Zhao, Xiaodong Li, Xiaokun Zhang, Shiming Ge, Dongqing
  Zou, Bin Zhou, and Xinghui Zhou.
\newblock Aesthetic attributes assessment of images.
\newblock In \emph{International Conference on Multimedia}, pages 311--319.
  ACM, 2019.

\bibitem[Kang et~al.(2020)Kang, Valenzise, and Dufaux]{kang2020eva}
Chen Kang, Giuseppe Valenzise, and Fr{\'e}d{\'e}ric Dufaux.
\newblock Eva: An explainable visual aesthetics dataset.
\newblock In \emph{Joint Workshop on Aesthetic and Technical Quality Assessment
  of Multimedia and Media Analytics for Societal Trends}, pages 5--13, 2020.

\bibitem[Ke et~al.(2021)Ke, Wang, Wang, Milanfar, and Yang]{ke2021musiq}
Junjie Ke, Qifei Wang, Yilin Wang, Peyman Milanfar, and Feng Yang.
\newblock Musiq: Multi-scale image quality transformer.
\newblock In \emph{International Conference on Computer Vision}, pages
  5148--5157. IEEE/CVF, 2021.

\bibitem[Kong et~al.(2016)Kong, Shen, Lin, Mech, and Fowlkes]{kong2016photo}
Shu Kong, Xiaohui Shen, Zhe Lin, Radomir Mech, and Charless Fowlkes.
\newblock Photo aesthetics ranking network with attributes and content
  adaptation.
\newblock In \emph{European Conference on Computer Vision}, pages 662--679.
  Springer, 2016.

\bibitem[Lee et~al.(2018)Lee, Kim, Lee, and Kim]{lee2018photographic}
Jun-Tae Lee, Han-Ul Kim, Chul Lee, and Chang-Su Kim.
\newblock Photographic composition classification and dominant geometric
  element detection for outdoor scenes.
\newblock \emph{Journal of Visual Communication and Image Representation},
  55:\penalty0 91--105, 2018.

\bibitem[Li et~al.(2022)Li, Li, Xiong, and Hoi]{li2022blip}
Junnan Li, Dongxu Li, Caiming Xiong, and Steven Hoi.
\newblock Blip: Bootstrapping language-image pre-training for unified
  vision-language understanding and generation.
\newblock In \emph{International Conference on Machine Learning}, 2022.

\bibitem[Liu et~al.(2019)Liu, Ott, Goyal, Du, Joshi, Chen, Levy, Lewis,
  Zettlemoyer, and Stoyanov]{DBLP:journals/corr/abs-1907-11692}
Yinhan Liu, Myle Ott, Naman Goyal, Jingfei Du, Mandar Joshi, Danqi Chen, Omer
  Levy, Mike Lewis, Luke Zettlemoyer, and Veselin Stoyanov.
\newblock Roberta: {A} robustly optimized {BERT} pretraining approach.
\newblock \emph{CoRR}, abs/1907.11692, 2019.
\newblock URL \url{http://arxiv.org/abs/1907.11692}.

\bibitem[Lu et~al.(2014)Lu, Lin, Jin, Yang, and Wang]{lu2014rapid}
Xin Lu, Zhe Lin, Hailin Jin, Jianchao Yang, and James~Z Wang.
\newblock Rapid: Rating pictorial aesthetics using deep learning.
\newblock In \emph{International Conference on Multimedia}, pages 457--466.
  ACM, 2014.

\bibitem[Luo et~al.(2011)Luo, Wang, and Tang]{cuhk-pq}
Wei Luo, Xiaogang Wang, and Xiaoou Tang.
\newblock Content-based photo quality assessment.
\newblock In \emph{International Conference on Computer Vision}, volume~15,
  pages 2206--2213, 11 2011.

\bibitem[Ma et~al.(2017)Ma, Liu, and Wen~Chen]{ma2017lamp}
Shuang Ma, Jing Liu, and Chang Wen~Chen.
\newblock A-lamp: Adaptive layout-aware multi-patch deep convolutional neural
  network for photo aesthetic assessment.
\newblock In \emph{Conference on Computer Vision and Pattern Recognition},
  pages 4535--4544. IEEE, 2017.

\bibitem[Marchesotti et~al.(2011)Marchesotti, Perronnin, Larlus, and
  Csurka]{marchesotti2011assessing}
Luca Marchesotti, Florent Perronnin, Diane Larlus, and Gabriela Csurka.
\newblock Assessing the aesthetic quality of photographs using generic image
  descriptors.
\newblock In \emph{International Conference on Computer Vision}, pages
  1784--1791. IEEE, 2011.

\bibitem[Murray et~al.(2012)Murray, Marchesotti, and Perronnin]{murray2012ava}
Naila Murray, Luca Marchesotti, and Florent Perronnin.
\newblock Ava: A large-scale database for aesthetic visual analysis.
\newblock In \emph{Conference on Computer Vision and Pattern Recognition},
  pages 2408--2415. IEEE, 2012.

\bibitem[Pandey et~al.(2017)Pandey, Rajpoot, and Saraswat]{pandey2017twitter}
Avinash~Chandra Pandey, Dharmveer~Singh Rajpoot, and Mukesh Saraswat.
\newblock Twitter sentiment analysis using hybrid cuckoo search method.
\newblock \emph{Information Processing \& Management}, 53\penalty0
  (4):\penalty0 764--779, 2017.

\bibitem[Pedregosa et~al.(2011)Pedregosa, Varoquaux, Gramfort, Michel, Thirion,
  Grisel, Blondel, Prettenhofer, Weiss, Dubourg, Vanderplas, Passos,
  Cournapeau, Brucher, Perrot, and Duchesnay]{scikit-learn}
F.~Pedregosa, G.~Varoquaux, A.~Gramfort, V.~Michel, B.~Thirion, O.~Grisel,
  M.~Blondel, P.~Prettenhofer, R.~Weiss, V.~Dubourg, J.~Vanderplas, A.~Passos,
  D.~Cournapeau, M.~Brucher, M.~Perrot, and E.~Duchesnay.
\newblock Scikit-learn: Machine learning in {P}ython.
\newblock \emph{Journal of Machine Learning Research}, 12:\penalty0 2825--2830,
  2011.

\bibitem[Punith and Raketla(2021)]{punith2021sentiment}
NS~Punith and Krishna Raketla.
\newblock Sentiment analysis of drug reviews using transfer learning.
\newblock In \emph{2021 Third International Conference on Inventive Research in
  Computing Applications (ICIRCA)}, pages 1794--1799. IEEE, 2021.

\bibitem[Rao et~al.(2020)Rao, Wang, Xu, Jiang, Huang, Zhou, and
  Lin]{rao2020unified}
Anyi Rao, Jiaze Wang, Linning Xu, Xuekun Jiang, Qingqiu Huang, Bolei Zhou, and
  Dahua Lin.
\newblock A unified framework for shot type classification based on subject
  centric lens.
\newblock In \emph{European Conference on Computer Vision}, pages 17--34.
  Springer, 2020.

\bibitem[Rocklage et~al.(2021)Rocklage, Rucker, and Nordgren]{rocklage2021mass}
Matthew~D Rocklage, Derek~D Rucker, and Loran~F Nordgren.
\newblock Mass-scale emotionality reveals human behaviour and marketplace
  success.
\newblock \emph{Nature human behaviour}, 5\penalty0 (10):\penalty0 1323--1329,
  2021.

\bibitem[Rosenthal et~al.(2017)Rosenthal, Farra, and
  Nakov]{rosenthal-etal-2017-semeval}
Sara Rosenthal, Noura Farra, and Preslav Nakov.
\newblock {S}em{E}val-2017 task 4: Sentiment analysis in {T}witter.
\newblock In \emph{International Workshop on Semantic Evaluation
  ({S}em{E}val-2017)}, pages 502--518, Vancouver, Canada, 2017. Association for
  Computational Linguistics.

\bibitem[Salas-Z{\'a}rate et~al.(2017)Salas-Z{\'a}rate, Medina-Moreira,
  Lagos-Ortiz, Luna-Aveiga, Rodriguez-Garcia, and
  Valencia-Garcia]{salas2017sentiment}
Mar{\'\i}a del~Pilar Salas-Z{\'a}rate, Jose Medina-Moreira, Katty Lagos-Ortiz,
  Harry Luna-Aveiga, Miguel~Angel Rodriguez-Garcia, and Rafael Valencia-Garcia.
\newblock Sentiment analysis on tweets about diabetes: an aspect-level
  approach.
\newblock \emph{Computational and mathematical methods in medicine}, 2017,
  2017.

\bibitem[Schifanella et~al.(2015)Schifanella, Redi, and
  Aiello]{schifanella2015image}
Rossano Schifanella, Miriam Redi, and Luca~Maria Aiello.
\newblock An image is worth more than a thousand favorites: Surfacing the
  hidden beauty of flickr pictures.
\newblock In \emph{International AAAI Conference on Web and Social Media},
  volume~9, pages 397--406, 2015.

\bibitem[Talebi and Milanfar(2018)]{talebi2018nima}
Hossein Talebi and Peyman Milanfar.
\newblock Nima: Neural image assessment.
\newblock \emph{IEEE Transactions on Image Processing}, 27\penalty0
  (8):\penalty0 3998--4011, 2018.

\bibitem[Wang et~al.(2019)Wang, Yang, Zhang, and Zhang]{wang2019neural}
Wenshan Wang, Su~Yang, Weishan Zhang, and Jiulong Zhang.
\newblock Neural aesthetic image reviewer.
\newblock \emph{IET Comput. Vis.}, 13\penalty0 (8):\penalty0 749--758, 2019.

\bibitem[Wolf et~al.(2020)Wolf, Debut, Sanh, Chaumond, Delangue, Moi, Cistac,
  Rault, Louf, Funtowicz, Davison, Shleifer, von Platen, Ma, Jernite, Plu, Xu,
  Le~Scao, Gugger, Drame, Lhoest, and
  Rush]{https://doi.org/10.48550/arxiv.1910.03771}
Thomas Wolf, Lysandre Debut, Victor Sanh, Julien Chaumond, Clement Delangue,
  Anthony Moi, Pierric Cistac, Tim Rault, Remi Louf, Morgan Funtowicz, Joe
  Davison, Sam Shleifer, Patrick von Platen, Clara Ma, Yacine Jernite, Julien
  Plu, Canwen Xu, Teven Le~Scao, Sylvain Gugger, Mariama Drame, Quentin Lhoest,
  and Alexander Rush.
\newblock Transformers: State-of-the-art natural language processing.
\newblock In \emph{Conference on Empirical Methods in Natural Language
  Processing: System Demonstrations}, pages 38--45. Association for
  Computational Linguistics, October 2020.

\bibitem[Xu et~al.(2020)Xu, Zhang, Wei, Sang, Li, and Yuan]{xu2020deep}
Yifei Xu, Nuo Zhang, Pingping Wei, Genan Sang, Li~Li, and Feng Yuan.
\newblock Deep neural framework with visual attention and global context for
  predicting image aesthetics.
\newblock \emph{IEEE Access}, 2020.

\bibitem[Yeo et~al.(2021)Yeo, See, Wong, and Goh]{yeo2021generating}
Yong-Yaw Yeo, John See, Lai-Kuan Wong, and Hui-Ngo Goh.
\newblock Generating aesthetic based critique for photographs.
\newblock In \emph{International Conference on Image Processing}, pages
  2523--2527. IEEE, 2021.

\bibitem[Zhang et~al.(2021)Zhang, Miao, and Yu]{zhang2021comprehensive}
Jiajing Zhang, Yongwei Miao, and Jinhui Yu.
\newblock A comprehensive survey on computational aesthetic evaluation of
  visual art images: Metrics and challenges.
\newblock \emph{IEEE Access}, 2021.

\bibitem[Zhang et~al.(2014)Zhang, Gao, Zimmermann, Tian, and
  Li]{zhang2014fusion}
Luming Zhang, Yue Gao, Roger Zimmermann, Qi~Tian, and Xuelong Li.
\newblock Fusion of multichannel local and global structural cues for photo
  aesthetics evaluation.
\newblock \emph{IEEE Transactions on Image Processing}, 23\penalty0
  (3):\penalty0 1419--1429, 2014.

\bibitem[Zhang and Zheng(2016)]{zhang2016comparison}
Xueying Zhang and Xianghan Zheng.
\newblock Comparison of text sentiment analysis based on machine learning.
\newblock In \emph{International Symposium on Parallel and Distributed
  Computing (ISPDC)}, pages 230--233. IEEE, 2016.

\bibitem[Zhou et~al.(2016)Zhou, Lu, Zhang, and Wang]{zhou2016joint}
Ye~Zhou, Xin Lu, Junping Zhang, and James~Z Wang.
\newblock Joint image and text representation for aesthetics analysis.
\newblock In \emph{International Conference on Multimedia}, pages 262--266.
  ACM, 2016.

\end{thebibliography}
}

\clearpage

\section*{Checklist}


\begin{enumerate}

\item For all authors...
\begin{enumerate}
  \item Do the main claims made in the abstract and introduction accurately reflect the paper's contributions and scope?
    \answerYes{The main claims are listed at the end of Section~\ref{sec:Intro}.}
  \item Did you describe the limitations of your work?
    \answerYes{The limitations are described in Section~\ref{sec:discussion}.}
  \item Did you discuss any potential negative societal impacts of your work?
    \answerYes{The potential negative societal impacts are described in Section~\ref{sec:discussion}.}
  \item Have you read the ethics review guidelines and ensured that your paper conforms to them?
    \answerYes{We additionally comment on different ethics consideration on Appendix~\ref{sec:ethics}.}
\end{enumerate}

\item If you are including theoretical results...
\begin{enumerate}
  \item Did you state the full set of assumptions of all theoretical results?
    \answerNA{}
	\item Did you include complete proofs of all theoretical results?
    \answerNA{}{}
\end{enumerate}

\item If you ran experiments (e.g. for benchmarks)...
\begin{enumerate}
  \item Did you include the code, data, and instructions needed to reproduce the main experimental results (either in the supplemental material or as a URL)?
    \answerYes{All necessary code is available on this repository \url{https://github.com/mediatechnologycenter/aestheval}.}
  \item Did you specify all the training details (e.g., data splits, hyperparameters, how they were chosen)?
    \answerYes{See Appendix~\ref{sec:implementation}.}
	\item Did you report error bars (e.g., with respect to the random seed after running experiments multiple times)?
    \answerNo{Training are very compute intensive. We can only run the training once per experiment using a random seed (arbitrarily selected).}
	\item Did you include the total amount of compute and the type of resources used (e.g., type of GPUs, internal cluster, or cloud provider)?
    \answerYes{See Appendix~\ref{sec:resources}}
\end{enumerate}

\item If you are using existing assets (e.g., code, data, models) or curating/releasing new assets...
\begin{enumerate}
  \item If your work uses existing assets, did you cite the creators?
    \answerYes{}
  \item Did you mention the license of the assets?
    \answerYes{We describe the license of used source of data in Appendix~\ref{sec:License}}
  \item Did you include any new assets either in the supplemental material or as a URL?
    \answerYes{}
  \item Did you discuss whether and how consent was obtained from people whose data you're using/curating?
    \answerYes{See Appendix~\ref{sec:ethics}}
  \item Did you discuss whether the data you are using/curating contains personally identifiable information or offensive content?
    \answerYes{See Appendix~\ref{sec:ethics}}
\end{enumerate}

\item If you used crowdsourcing or conducted research with human subjects...
\begin{enumerate}
  \item Did you include the full text of instructions given to participants and screenshots, if applicable?
    \answerNA{}
  \item Did you describe any potential participant risks, with links to Institutional Review Board (IRB) approvals, if applicable?
    \answerNA{}
  \item Did you include the estimated hourly wage paid to participants and the total amount spent on participant compensation?
    \answerNA{}
\end{enumerate}

\end{enumerate}


\clearpage

\appendix

\section{Dataset Analysis}
\label{app:datanalysis}
\subsection{Subreddits Analysis}
\label{app:subanaly}
\begin{figure}[ht!]
    \centering
    \includegraphics[width=.7\linewidth]{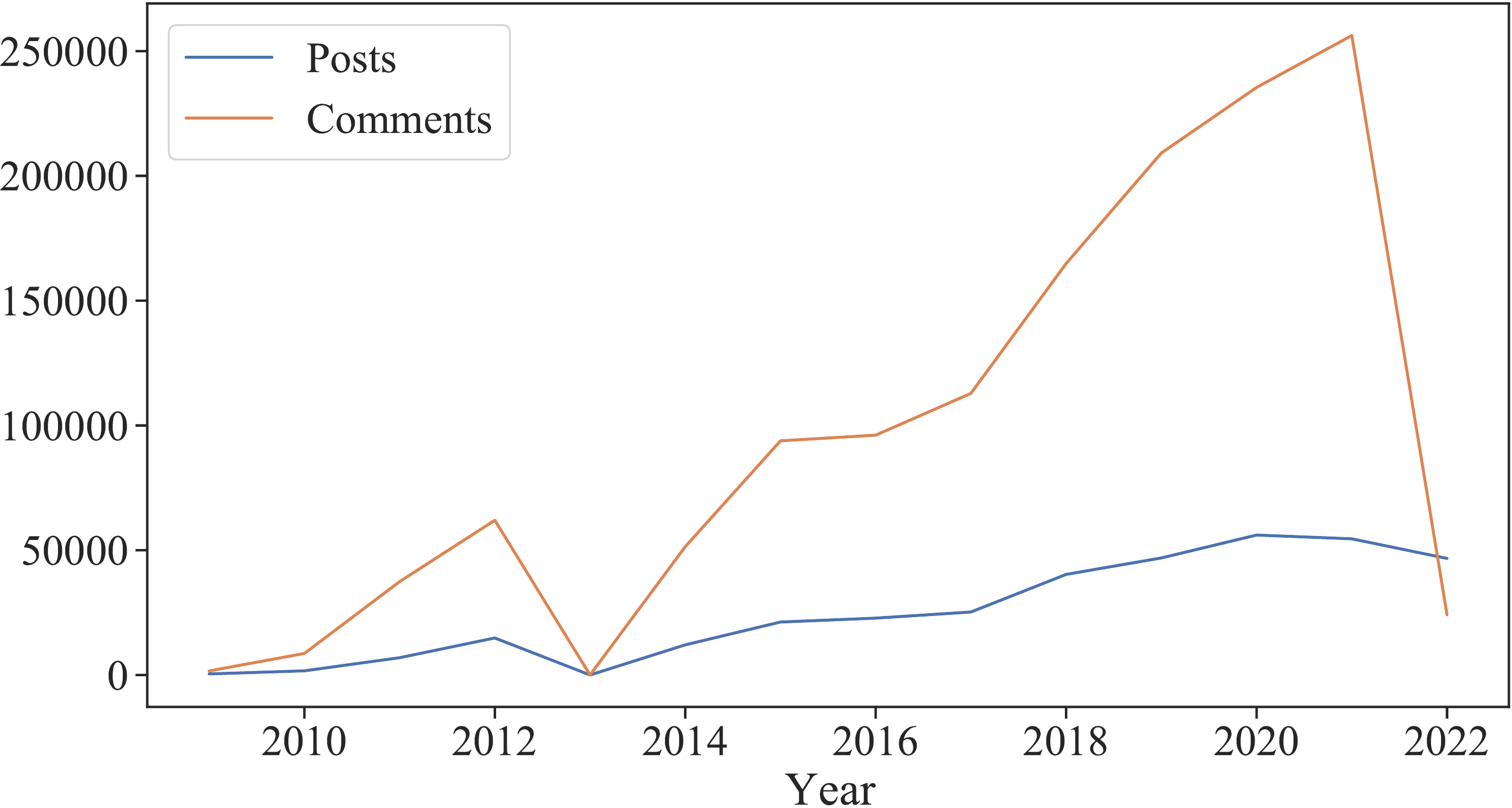}
    \caption{The number of posts and comments between May 2009 and February 2022 for the 6 considered subreddits.}
    \label{fig:posts-comments}
\end{figure}

\ADD{There are 1.5M members in the \texttt{/r/photocritique} subreddit. Since it is not possible to collect demographic information about subreddit members, we report the statistic related to a recent analysis about Reddit \footnote{\url{https://www.statista.com/topics/5672/reddit/\#topicHeader} (Accessed on 22/08/2022)}. Slight majority of Reddit users are male (61\%). 48\% of Reddit users are in the US, followed by the UK, Canada, Australia and Germany. People between the age of 18 and 29 make up Reddit's largest user base (64\%). The second biggest age group is 30 to 49 (29\%). Teenagers below 15 are not very active on Reddit. Only 7\% of Reddit users are over 50.

In light of the previous statistics, it is necessary to underline that the data treated in our dataset, therefore the inferred concept of aesthetics, presents a bias due to the limited cultural and geographical integration of the people who produced the information.

Here, it follows a deeper analysis of \texttt{/r/photocritique} subreddit.} Figure \ref{fig:posts-comments} shows the number of posts and comments per year downloaded from the seven subreddits we have selected. We observe that the number of posts and comments increase over the course of time. Data from 2013 could not be retrieved due to problems with Pushshift~\footnote{\url{https://www.reddit.com/r/pushshift/comments/sb982i/very_recent_data_missing/}}. Since 2015 there have been a number of posts over 20K and a number of comments that exceeds 100K until reaching the peak of 250K in 2021. Furthermore, although the posts are substantially fewer than the comments, the posts have reached a constant level of over 50,000 per year.
%

\subsection{Image Resolution} 
\label{app:imageres}
\begin{figure}[ht]
    \centering
    \includegraphics[width=.8\linewidth]{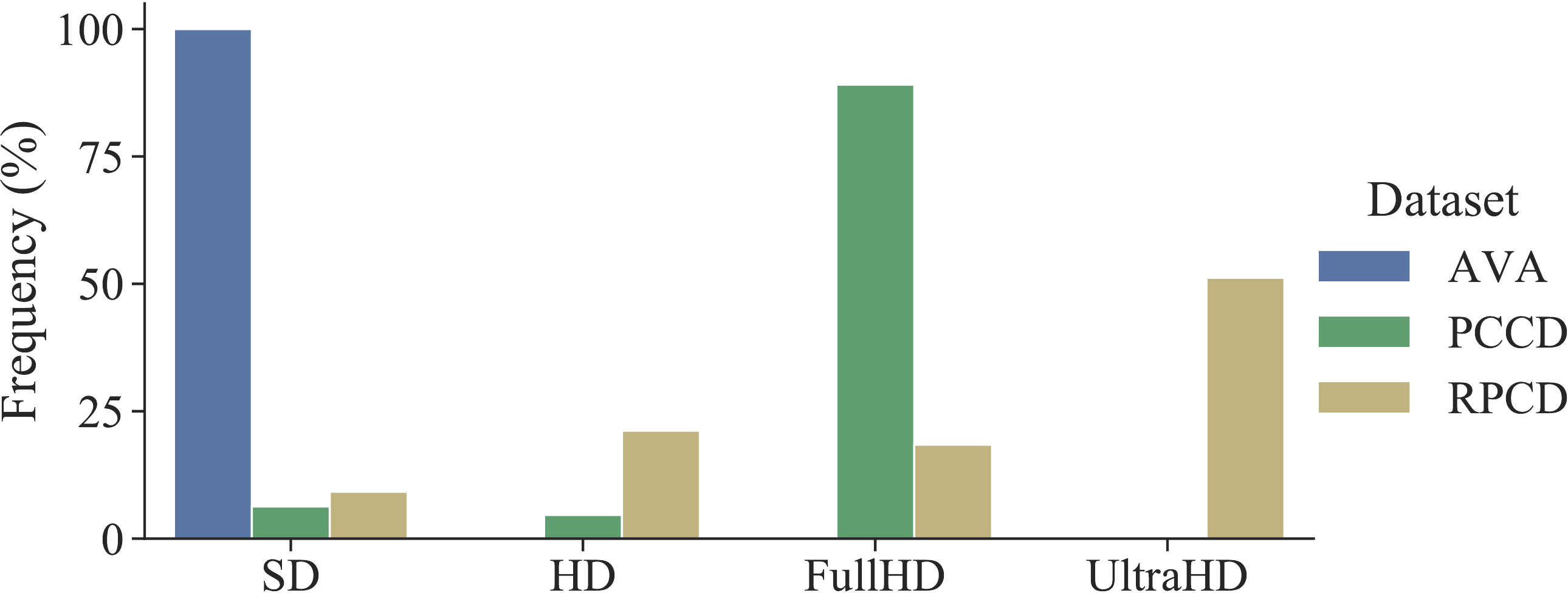}
    \caption{Image distribution of AVA, PCCD and our dataset, RPCD, for various standard image resolutions.}
    \label{fig:im-resolutions}
\end{figure}

We investigate the resolution of the images of three datasets, namely AVA, PCCD, and our RPCD. We categorize images into 4 common image resolutions in still camera photography, namely Standard Definition -- SD ($720 \times 576$ pixels), High Definition -- HD ($1280 \times 720$ pixels), FullHD ($1920 \times 1080$ pixels), and UltraHD ($3840 \times 2160$ pixels). In Figure \ref{fig:im-resolutions} the distributions of the images for the three datasets are plotted with respect to the four considered resolutions. As it is possible to see, our dataset it the only one that has UltraHD images. Most of the images are UltraHD resolution (51.20\%), but there are also images for the other three resolutions. On the other hand, all AVA images have a resolution of $720 \times 576$ pixels, while most PCCD images (i.e. 89.07\%) have FullHD resolution.

\subsection{Sentiment Polarity Classification}
\label{appx:sentiment}
\begin{figure}[ht]
    \centering
    \includegraphics[width=.7\linewidth]{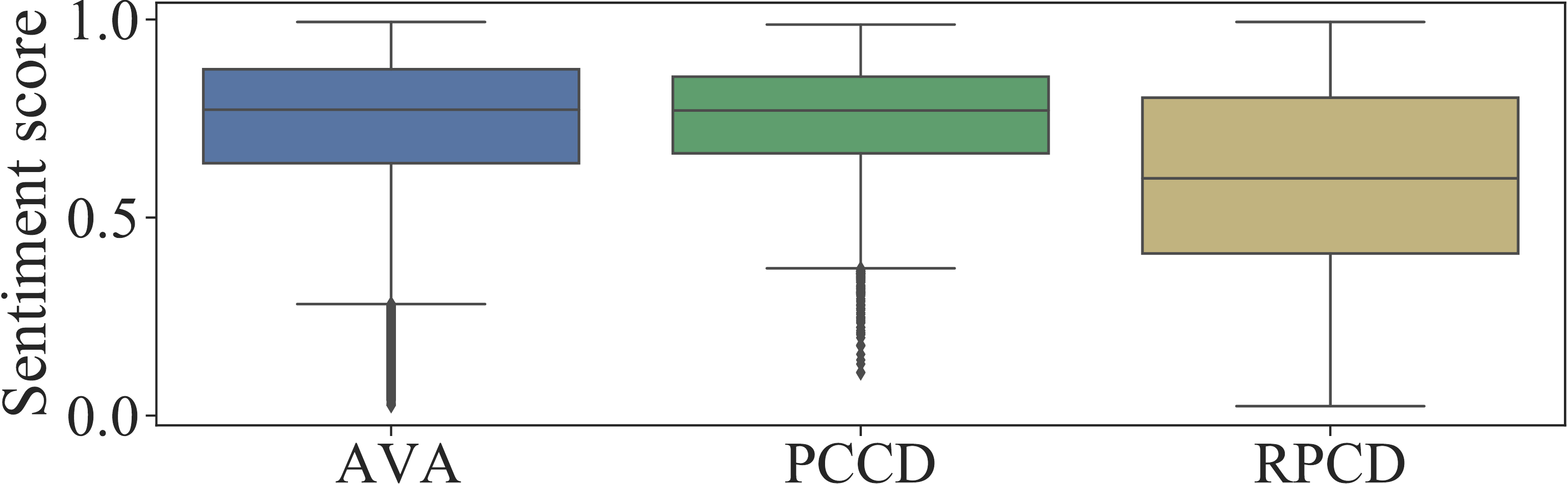}
    \caption{Boxplots of sentiment score distributions for the three considered datasets, namely AVA, PCCD and our RPCD.}
    \label{fig:boxplot-sentiment-score}
\end{figure}

We delve into the analysis of the sentiment score distributions of our dataset and those of AVA and PCCD. Figure \ref{fig:boxplot-sentiment-score} shows the spreads of the sentiment scores for the three datasets. AVA and PCCD have very similar median and standard deviation, namely 0.77 and about 0.15. Our RPCD on the other hand has a median of 0.60 and a larger standard deviation (i.e., 0.25). This difference between ours and the other datasets indicates that RPCD have a richer representation of the whole aesthetic taste spectrum, providing information about why an image have a specific score for high and low sentiment scores.
%

\begin{figure}
    \centering
    \includegraphics[width=\linewidth]{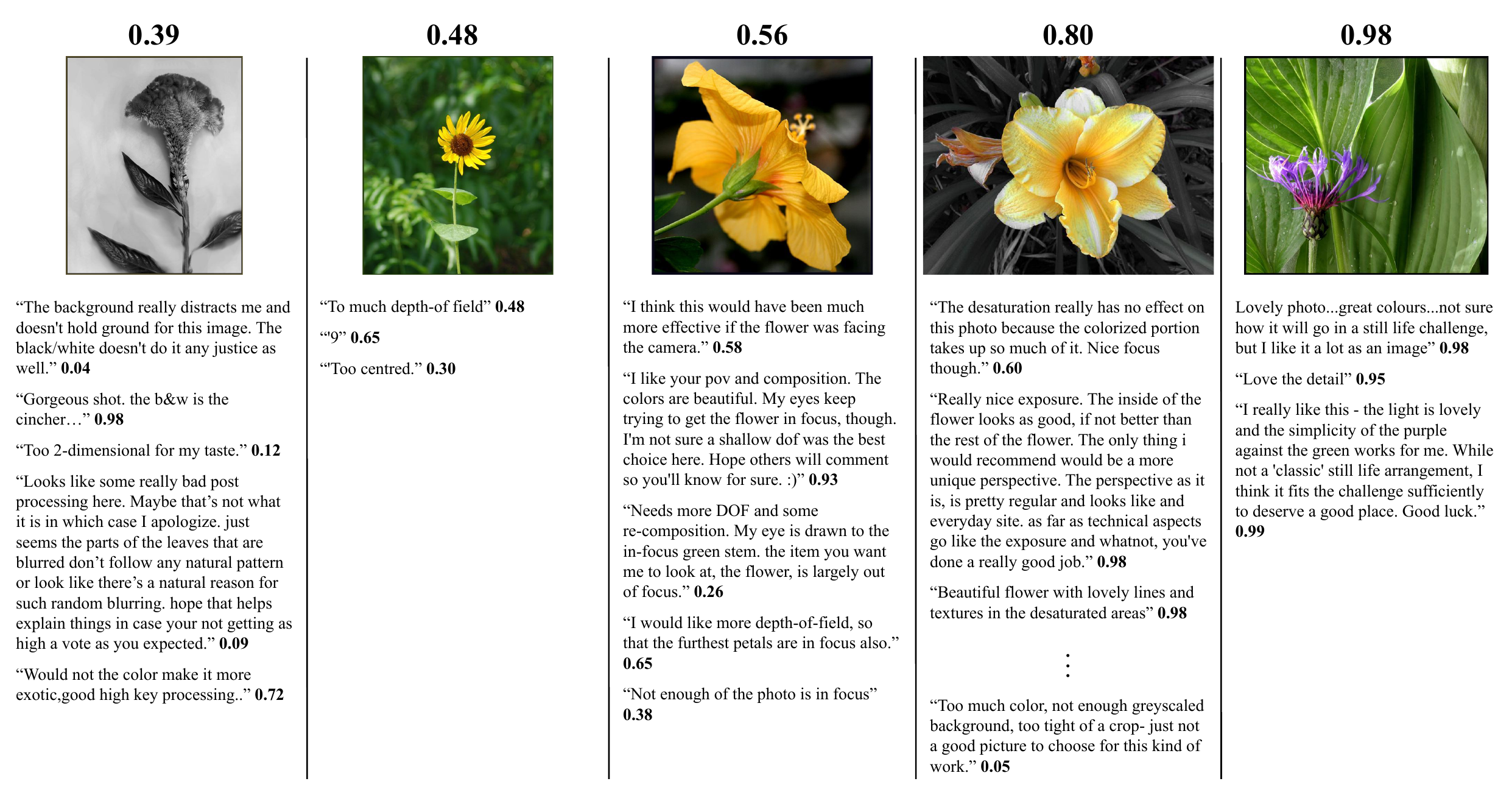}
    \caption{AVA samples annotated with an aesthetic score of 5, whose sentiment score we propose varies between 0.39 and 0.98. For each image we report the overall sentiment score (top of the image) and comments with the corresponding predicted sentiment score in bold.}
    \label{fig:avascore5vssentimentscore}
\end{figure}

%
Figure \ref{fig:avascore5vssentimentscore} reports some samples of the AVA dataset whose aesthetic score given by the human raters is equal to 5 (i.e., average score of the distribution), while our sentiment score span almost the entire range. It can be seen that the comments concern different aspects of photography. For example, for the central image a user has concerns about the pose of the subject ``I think this would have been much more effective if the flower was facing the camera.'', while another user would have preferred a different optical technique, i.e., ``I would like more depth-of-field, so that the furthest petals are in focus also''. Sometimes there can be very conflicting opinions in the comments (see the first image on the left). In general, comments reveal many facets of judgment shaped by the polarity of sentiment. This therefore justifies the difference between the annotated aesthetic score and the estimated sentiment score.
\subsection{Content Analysis}
\label{appx:content}
We automatically analyze image content by using image classifiers for both semantic and composition aspects. In this section we detail the classifiers design and training and some qualitative results on our RPCD dataset.
\paragraph{Semantic Content and Composition Rule.} To categorize the semantic content and composition of RPCD images, we use two classifiers based on the same backbone, namely the Vision Transformer (ViT) presented in \cite{dosovitskiy2020image}. In particular, we use the ViT parameters learned on ImageNet (keeping them freezed on the new tasks). The last linear classification layer is peculiar to each task and its parameters are trained. We use the same hyperparameters for the two classifiers, that is SGD with momentum equal to 0.9 and weight decay of 1$e$-4. We train using batches of 32 images for 90 epochs with an initial learning rate of 0.01, that is then dropped every 30 epochs by a factor of 0.1.

\begin{figure}[ht]
    \centering
    \setlength{\tabcolsep}{1pt}
    \begin{tabular}{cccccc}
        \includegraphics[width=.15\textwidth,height=40px]{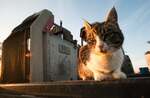} & \includegraphics[width=.15\textwidth,height=40px]{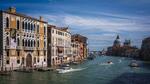} & \includegraphics[width=.13\textwidth,height=50px]{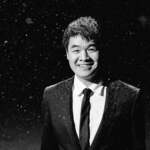} & \includegraphics[width=.15\textwidth,height=40px]{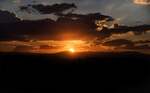} &
        \includegraphics[width=.15\textwidth,height=40px]{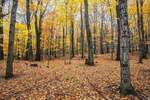} & 
        \includegraphics[width=.12\textwidth,height=55px]{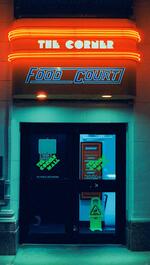} \\
        \includegraphics[width=.15\textwidth,height=40px]{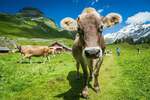} & \includegraphics[width=.15\textwidth,height=40px]{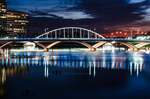} & \includegraphics[width=.15\textwidth,height=40px]{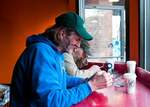} & \includegraphics[width=.15\textwidth,height=40px]{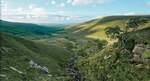} & \includegraphics[width=.15\textwidth,height=40px]{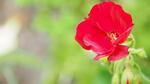} &  \includegraphics[width=.15\textwidth,height=40px]{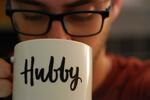} \\
        Animal & Architecture & Human & Landscape & Plant & Static
    \end{tabular}
    \caption{Images from our RCPD dataset categorized with respect to the semantic content.}
    \label{fig:semantic-class-samples}
\end{figure}

The semantic content classifier is trained to discriminate six different semantic content, namely \texttt{Animal}, \texttt{Architecture}, \texttt{Human}, \texttt{Landscape}, \texttt{Plant}, and \texttt{Static}. For this purpose, we use 15,981 images of the dataset CUHK-PQ \cite{cuhk-pq} (i.e. all the images of the dataset apart from those of the Night category). We split the whole dataset int 80\% training images and 20\% test images. The resulting classifier achieved an accuracy of 87.08\% on the test set. Figure \ref{fig:semantic-class-samples} reports two images from RPCD for each semantic category.

\begin{figure}[ht]
    \centering
    \setlength{\tabcolsep}{1pt}
    \begin{tabular}{ccccc}
        \includegraphics[width=.1\textwidth]{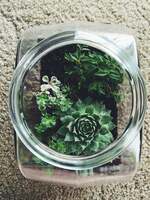} & \includegraphics[width=.1\textwidth,height=50px]{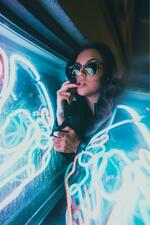} & \includegraphics[width=.1\textwidth,height=50px]{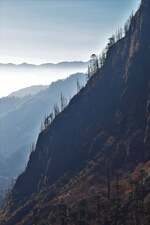} & \includegraphics[width=.1\textwidth,height=50px]{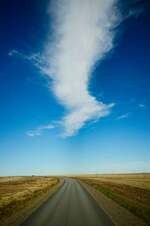}  &
        \includegraphics[width=.13\textwidth]{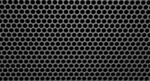} \\
        \includegraphics[width=.13\textwidth]{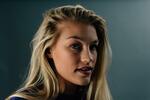} & \includegraphics[width=.13\textwidth]{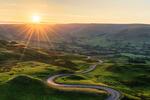} & \includegraphics[width=.13\textwidth]{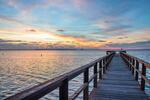} & \includegraphics[width=.13\textwidth]{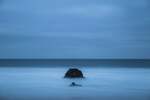} & \includegraphics[width=.13\textwidth,height=34px]{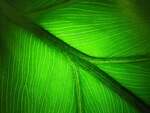} \\
        Center & Curved & Diagonal & Horizontal & Pattern \vspace{1em}\\
    \end{tabular}
    \setlength{\tabcolsep}{1pt}
    \begin{tabular}{cccc}
        \includegraphics[width=.13\textwidth]{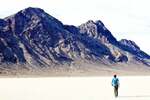} & 
        \includegraphics[width=.13\textwidth]{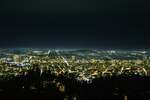} & 
        \includegraphics[width=.1\textwidth]{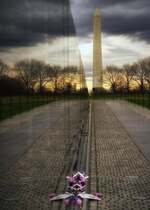} & 
        \includegraphics[width=.13\textwidth]{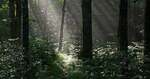} \\
        \includegraphics[width=.13\textwidth]{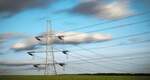} & \includegraphics[width=.13\textwidth]{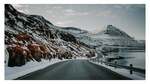} & 
        \includegraphics[width=.1\textwidth]{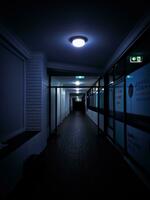} & \includegraphics[width=.1\textwidth]{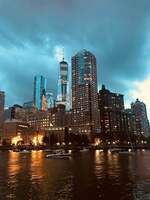} \\
        RoT & Symmetric & Triangle & Vertical
    \end{tabular}
    \caption{Images from our RCPD dataset categorized with respect to the main composition rule.}
    \label{fig:composition-class-samples}
\end{figure}

\begin{figure}[ht]
    \centering
    \setlength{\tabcolsep}{1pt}
    \begin{tabular}{ccccc}
        \includegraphics[width=.15\textwidth,height=40px]{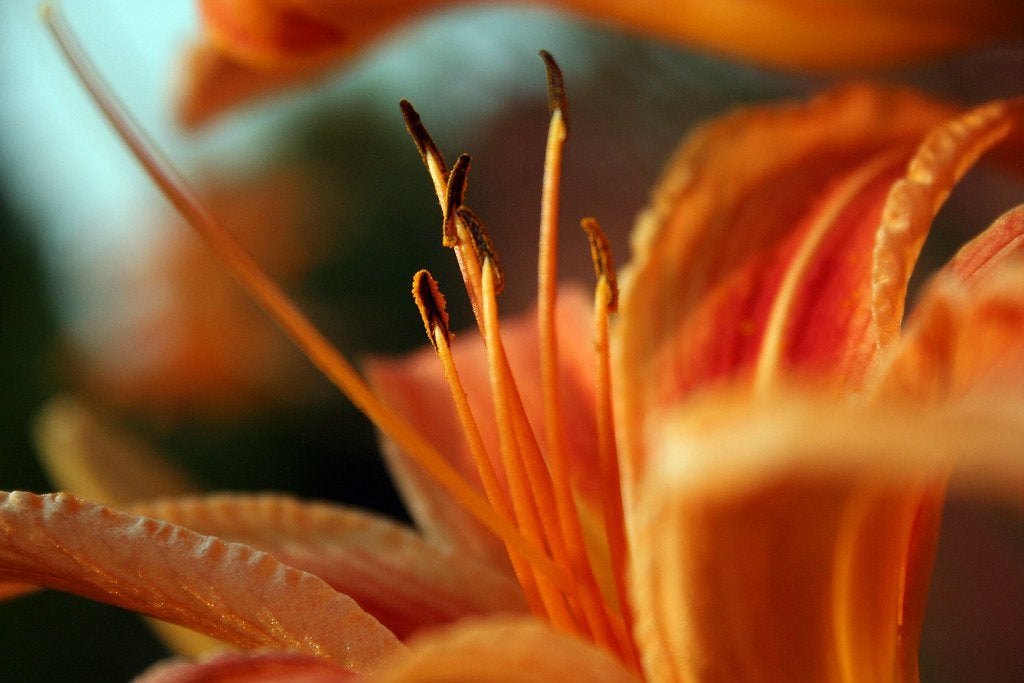} & \includegraphics[width=.15\textwidth,height=40px]{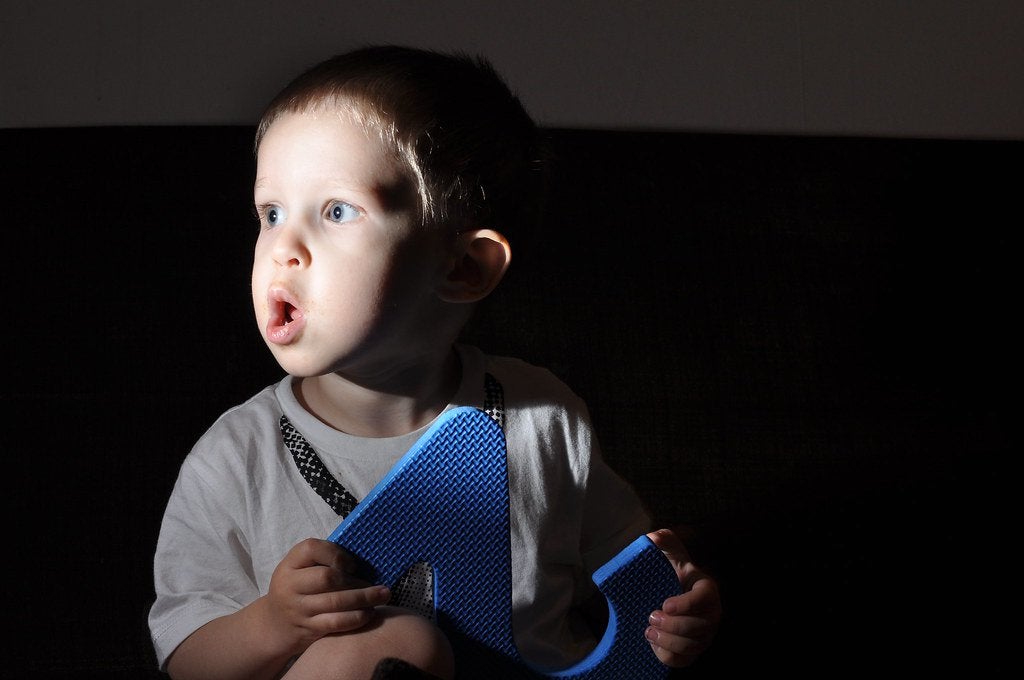} & \includegraphics[width=.15\textwidth,height=40px]{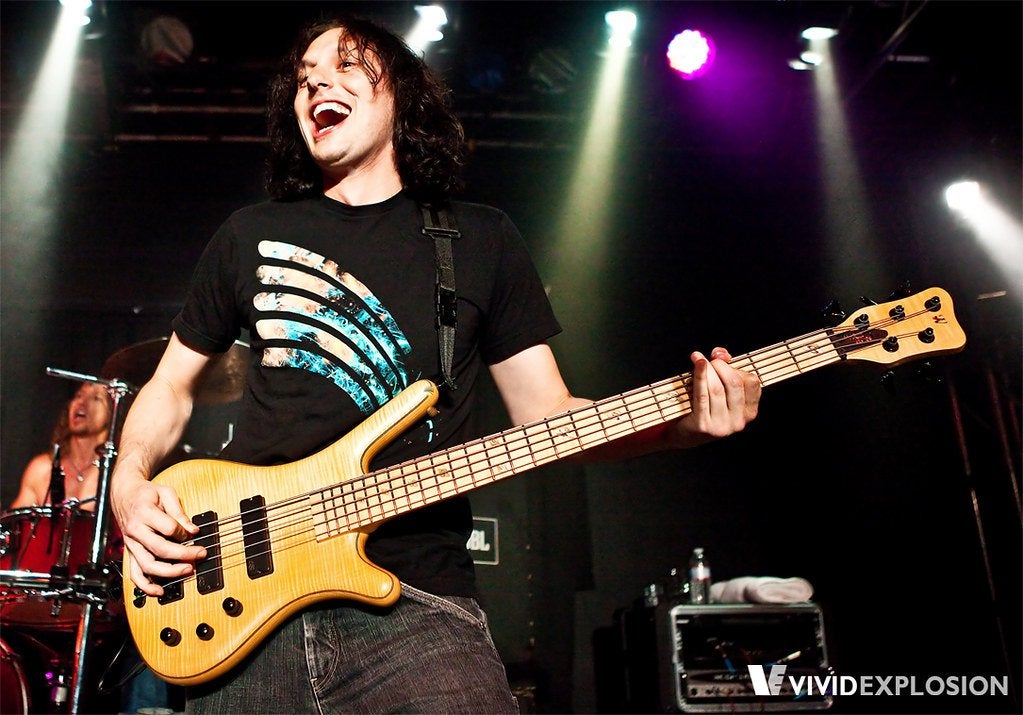} & \includegraphics[width=.15\textwidth,height=40px]{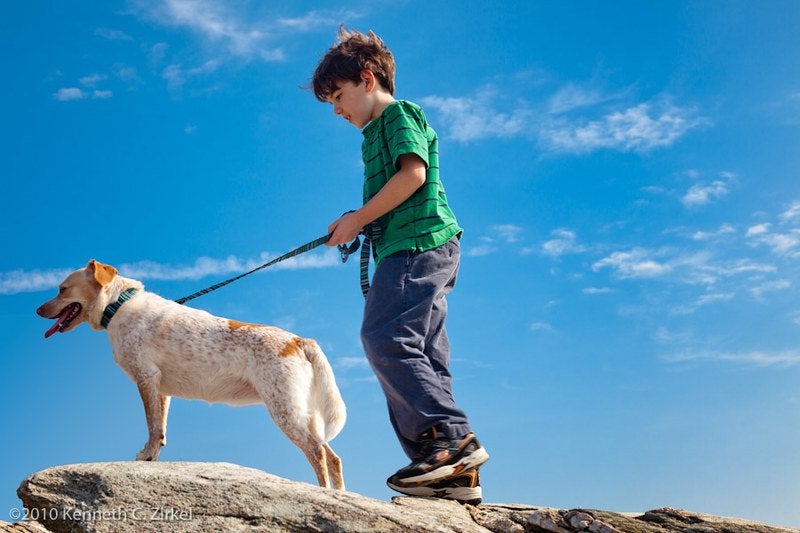} &
        \includegraphics[width=.15\textwidth,height=40px]{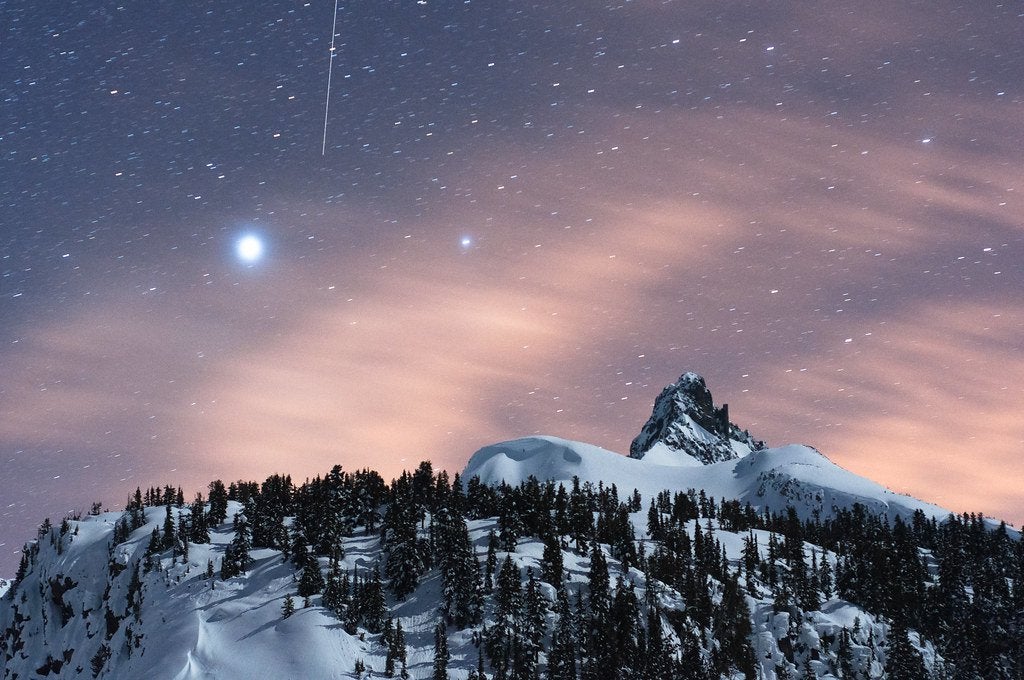} \\
        \includegraphics[width=.15\textwidth,height=40px]{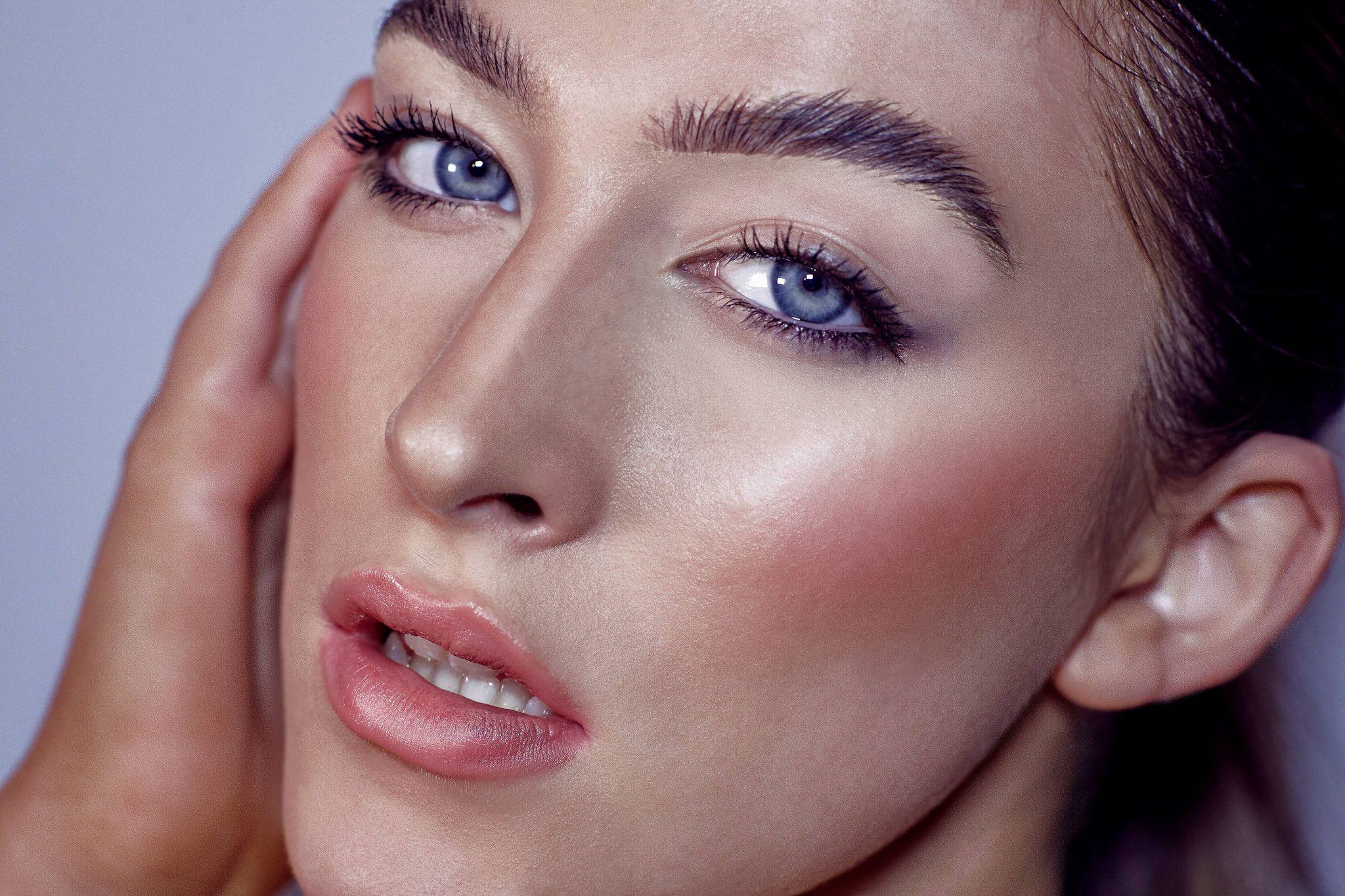} &
        \includegraphics[width=.15\textwidth,height=40px]{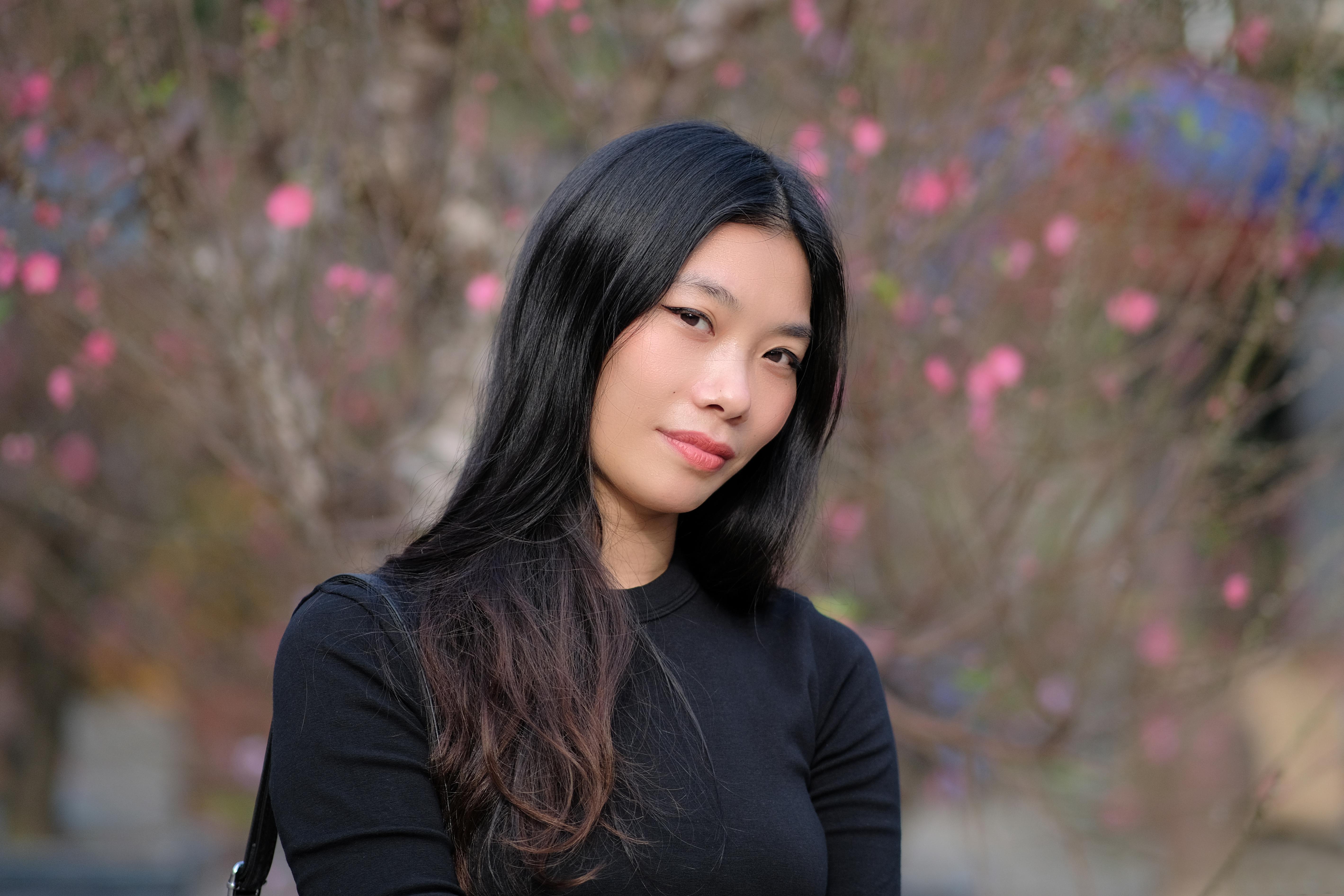} & \includegraphics[width=.15\textwidth,height=40px]{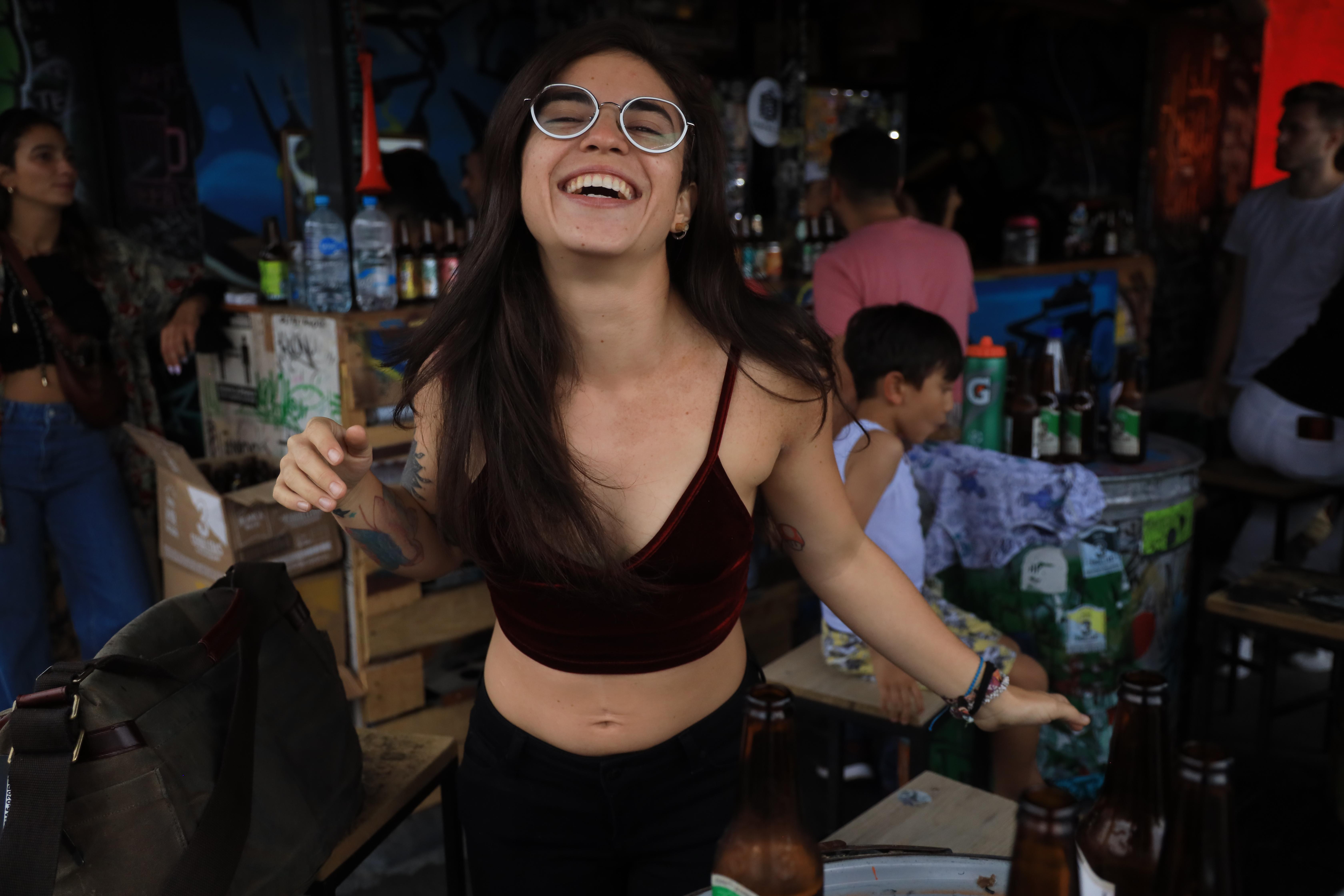} & \includegraphics[width=.15\textwidth,height=40px]{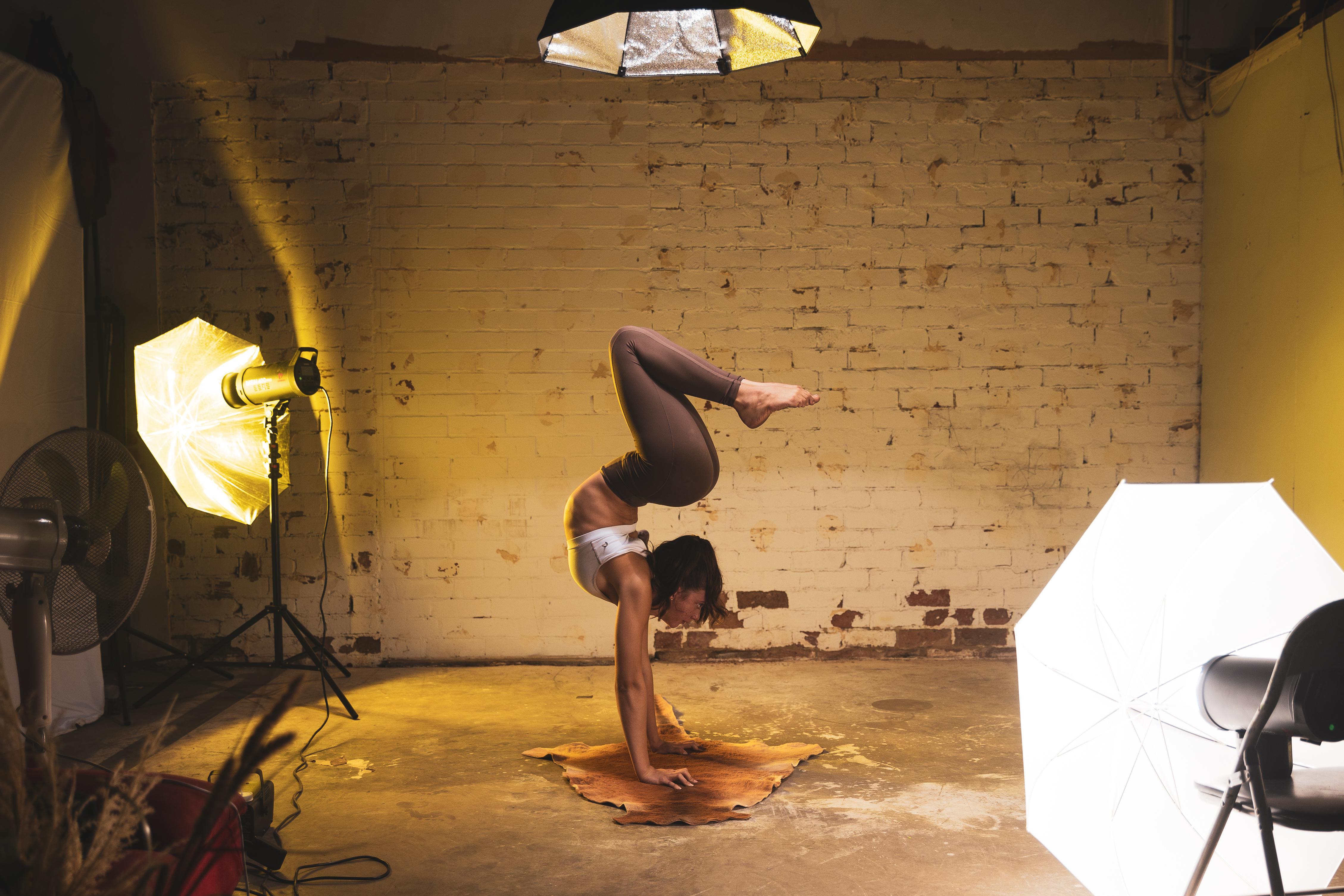} & \includegraphics[width=.15\textwidth,height=40px]{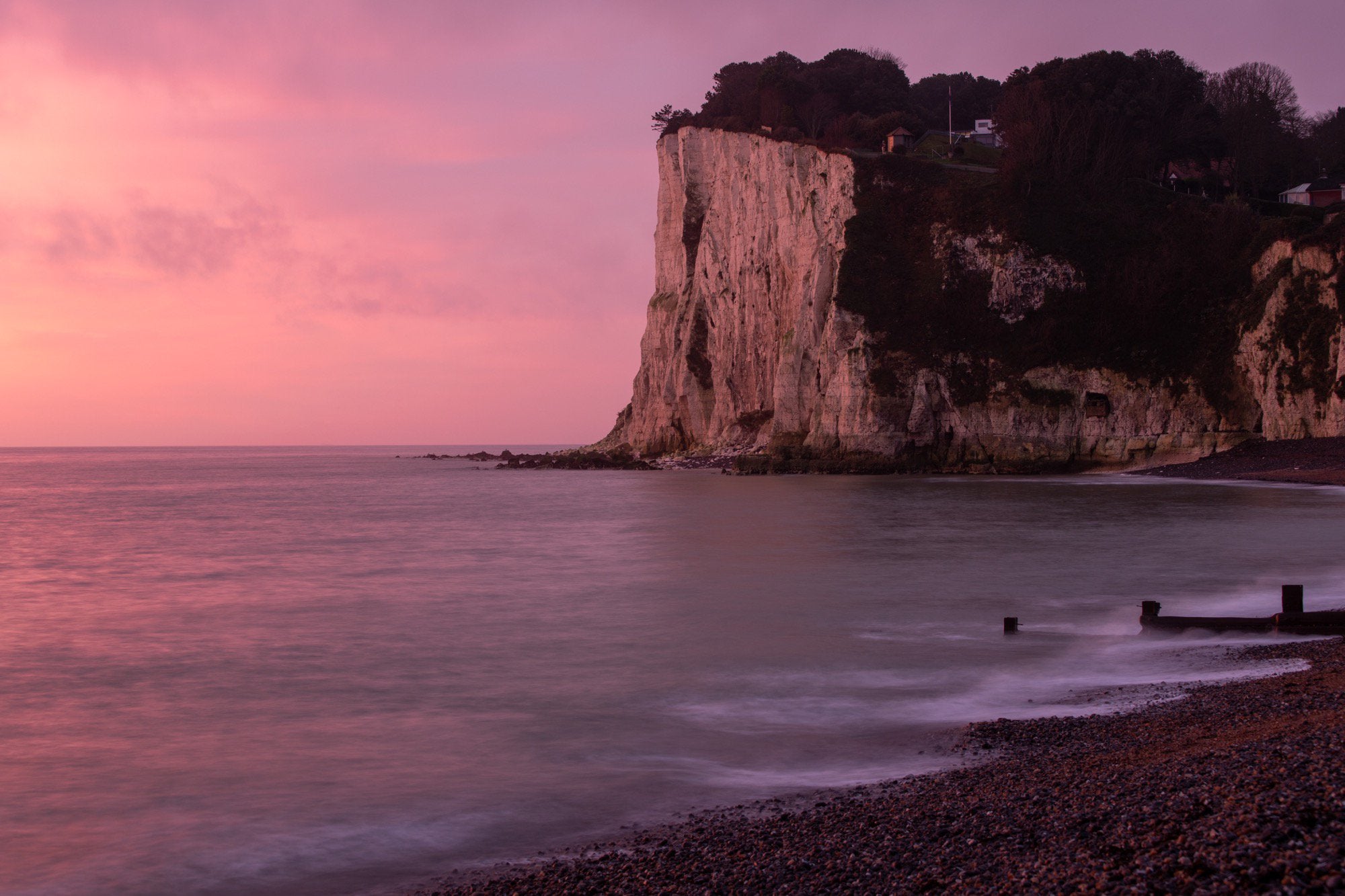} \\ \\
        ECS & CS & MS & LS & FS
    \end{tabular}
    \caption{Images from our RCPD dataset categorized with respect to the shot scale.}
    \label{fig:shotscale-class-samples}
\end{figure}

Our composition classifier is trained on the KU-PCP dataset \cite{lee2018photographic}, which consists of 4244 outdoor photographs. We exploit the data splits provided by the authors which comprise of a training set of 3169 images and 1075 validation images. Each image has been annotated by 18 human subject to categorize it into nine composition classes: \texttt{Center}, \texttt{Curved}, \texttt{Diagonal}, \texttt{Horizontal}, \texttt{Pattern}, \texttt{Rule of Thirds (RoT)}, \texttt{Symmetric}, \texttt{Triangle}, and \texttt{Vertical}. Since an image may follow multiple composition rules, each sample is given with one or more (at most 3) composition labels. Following \cite{hong2021composing}, images with more than one rule are trained multiple times for each ground-truth class. This training strategy is shown more effective than multi-label loss. The estimated accuracy on the test set is equal to 33.36\%. Figure \ref{fig:composition-class-samples} shows some images from the RPCD categorized for each composition rule.
\paragraph{Shot Scale.} We implement a Subject Guidance Network (SGNet) inspired by \cite{rao2020unified} to perform shot scale classification on images. The key idea is to use a subject map to determine the portion occupied by the subject with respect to the frame. We distinguish among five shot scale types, namely extreme close-up (ECS), close-up (CS), medium (MS), full (FS) and long (LS). The model is trained on the public MovieNet dataset \cite{huang2020movienet} and optimized with stochastic gradient descent using cross-entropy loss. We use a learning rate of 1$e$-3, batch size of 16 and we train for 60 epochs. We achieved 99.72$\%$ accuracy on the test set of the MovieNet dataset and observed a good generalization to the proposed RPCD dataset. The shot scale reveals information of how the photographer used the camera in order to emphasize either a location (long), an event (medium/full) or the identity of a subject (extreme close-up/close-up). Figure \ref{fig:shotscale-class-samples} reports some images annotated for each shot scale category.

\newpage
\paragraph{Aesthetic Aspect Prediction.} The aesthetic aspect of each comment is predicted using a transformer model, DistilBERT, implemented using HugginFace's transformers library~\cite{https://doi.org/10.48550/arxiv.1910.03771}. This approach differs with previous attempts of automatically labeling the aesthetic attributes of comments, which were based on keywords~\cite{jin2019aesthetic}. Instead, we fine-tune the language model for the text classification task of predicting the aesthetic attribute of a text using the PCCD~\cite{chang2017aesthetic} dataset, where 7 different classes are available: \texttt{general\_impression}, \texttt{subject\_of\_photo}, \texttt{composition}, \texttt{use\_of\_camera}, \texttt{depth\_of\_field}, \texttt{color\_lighting}, and \texttt{focus}. We use a learning rate of 2$e$-5, batch size of 16, weight decay of 0.01 and we train for 5 epochs. The rest of the parameters are left to the default ones in the HugginFace Trainer API. We randomly split the whole dataset in two folds: 90\% for training, and the remaining 10\% for validation and testing. Additionally, we clean URLs and escaped characters from the dataset. The fine-tuning converges at epoch 2, where the weighted metrics over the 7 different classes are: Precision, 0.8771; Recall, 0.8751; F1-score, 0.8755; and Accuracy, 0.8751.

\begin{figure}[ht!]
    \centering
    \includegraphics[width=.5\textwidth]{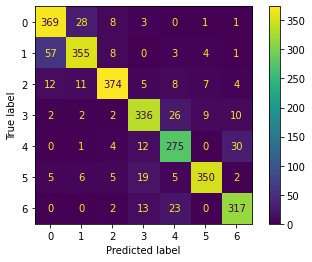}
    \caption{Correlation matrix of the aesthetic aspect classifier.}
    \label{fig:aesth_aspect_clf_corr_matrix}
\end{figure}

Figure~\ref{fig:aesth_aspect_clf_corr_matrix} shows the correlation matrix of the classifier performance on the test set. The classifier is available on HuggingFace's model hub~\footnote{\url{https://huggingface.co/daveni/aesthetic_attribute_classifier}}.

\ADD{\subsection{Explicit or offensive content}
\label{appx:offensive}
We use Detoxify~\footnote{\url{https://github.com/unitaryai/detoxify}}, a library to predict toxic comments, to carry out a preliminary analysis of the presence of offensive content in the dataset. We use the \textit{unbiased} model, a model that recognizes toxicity and minimizes this type of unintended bias with respect to mentions of identities (for example, minimize the bias towards the toxic class when a mention to a minority, which are often the target of offensive comments, is mentioned and the comment is not actually offensive). 

In Table~\ref{tab:detoxify} we show the results of this preliminary analysis of the presence of offensive content in the dataset using Detoxify to predict the "offensive probability" of the ~216K comments in the dataset. We have considered a comment to be offensive if the predictions probability for any of the labels is higher than 0.5. In total, there are ~8K comments with a predicted probability of being offensive greater that 0.5, which represent less than the 4\% of the total of comments in the dataset. 

\begin{table}[!ht]
    \centering
    \caption{Offensive content analysis of our RPCD using Detoxify.}
    \begin{tabular}{l|cc}
    \toprule
        Offensive label & Predicted Probability Mean & Total \\ \midrule
        toxicity & 2.889\% & 4369 \\
        severe\_toxicity & 0.019\% & 0 \\
        obscene  & 1.125\% & 2385 \\
        identity\_attack  & 0.374\% & 336 \\
        insult  & 0.672\% & 742 \\
        threat  & 0.418\% & 287 \\
        sexual\_explicit  & 0.259\% & 439 \\ \bottomrule
    \end{tabular}
    \label{tab:detoxify}
\end{table}
}
\subsection{Topic Modeling}
\label{appx:topic}

We use BERTopic~\cite{grootendorst2022bertopic} to clusterize the comments in all three datasets to compare the main topics being discussed. This method leverages on the document embeddings created using a text encoder 
to produce clusters after reducing the dimensionality of such embeddings. Then, TD-IDF is applied to the documents of the cluster to get the importance score of each word, obtaining the relevant topics in the cluster. 

To generate the topics, we used the automatic topic reduction feature available in the library to reduce the number of topics, starting from the least frequent topic, as long as it exceeds a minimum similarity of 0.915. We additionally sample 100K comments from AVA and Reddit datasets to avoid memory constraints. We describe the datasets topics in the Table \ref{tab:top30topics}. It shows the top 30 topics together with the count of documents belonging to each of them and the most important words per topic. Topics related to aesthetic attributes are in bold. We observe that in all of them we can find topics regarding aesthetic aspects such as composition, exposure, focus or color; but also topics related to the subject of the image such as sky, bird or flower.

\begin{table}[h]
\centering
\caption{Top 30 detected Topics on AVA, PCCD, and our RPCD.}
\label{tab:top30topics}
\resizebox{\linewidth}{!}{\begin{tabular}{rl|rl|rl}
\toprule
\multicolumn{2}{c|}{AVA} & \multicolumn{2}{c|}{PCCD} & \multicolumn{2}{c}{RPCD} \\
Count & Name &  Count & Name & Count & Name\\
\midrule
35798 &                 focus\_and\_challenge\_this &12585 &                           the\_and\_of\_to &43491 &                         and\_the\_to\_is \\
1417 &                         her\_she\_face\_shes &1256 &                        hi\_you\_work\_image &3441 &                      her\_she\_face\_hair \\
1170 &                flower\_flowers\_petals\_leaf &690 &               flower\_flowers\_petals\_rose &2770 &          horizon\_tree\_trees\_straighten \\
1150 &             \textbf{crop\_cropped\_tighter\_cropping} &641 &                        her\_eyes\_she\_face &2423 &                     bird\_dog\_cat\_birds \\
1081 &                         dog\_cat\_cats\_dogs &497 &             \textbf{exposure\_speed\_shutter\_water} &1971 &                  sky\_clouds\_cloud\_blue \\
904 &                    sky\_clouds\_cloud\_skies &469 &                  bird\_birds\_feathers\_the &1367 &            \textbf{crop\_cropped\_square\_tighter} \\
899 &               title\_titles\_without\_titled &451 &             \textbf{sharp\_focus\_looks\_resolution} &1218 &  building\_buildings\_tower\_architecture \\
810 &              ribbon\_red\_congrats\_deserved &434 &          \textbf{field\_depth\_shallow\_appropriate} &1217 &                        his\_him\_he\_face \\
786 &                tree\_trees\_branches\_branch &417 &        \textbf{subject\_interesting\_matter\_choice} &1064 &                 where\_taken\_live\_place \\
773 &                      water\_drops\_fog\_rain &372 &                \textbf{color\_lighting\_colors\_sky} &888 &            flower\_flowers\_petals\_focus \\
747 &          \textbf{composition\_composed\_shot\_nicely} &365 &                  tree\_trees\_branches\_the &821 &       \textbf{water\_reflection\_exposure\_puddle} \\
714 &           portrait\_self\_portraits\_candid &347 &            child\_baby\_children\_daughter &782 &                 boat\_boats\_ship\_water \\
687 &    reflection\_reflections\_mirror\_mirrors &326 &                 \textbf{iso\_noise\_speed\_shutter} &728 &       please\_titles\_examples\_specific \\
684 &              score\_averaged\_total\_autool &298 &      \textbf{perspective\_composition\_angle\_good }&675 &                  car\_cars\_truck\_front \\
653 &        comment\_done\_knowitall\_explaining &295 &              \textbf{dof\_diffraction\_focus\_good} &649 &               \textbf{iso\_shutter\_speed\_noise} \\
626 &               \textbf{bw\_conversion\_choice\_works} &257 & landscape\_location\_beautiful\_landscapes &617 &           \textbf{hdr\_range\_dynamic\_exposures} \\
620 &        \textbf{sharp\_sharpness\_sharper\_sharpened} &251 &                 \textbf{aperture\_fdepth\_field}&592 &          \textbf{bridge\_bridges\_leading\_lines} \\
617 &           \textbf{capture\_great\_wonderful\_colors} &225 &               \textbf{auto\_manual\_mode\_settings} &549 &  mountain\_mountains\_clouds\_foreground \\
597 &               \textbf{shadow\_shadows\_light\_harsh} &198 &                good\_very\_bad\_apparently &515 &        street\_road\_photography\_trails \\
564 &                   finish\_top\_congrats &181 &                     spot\_on\_looks\_seems &465 &              url\_thisurl\_oneurl\_heres \\
537 &                    bird\_birds\_beak\_eagle &178 &               \textbf{perfect\_looks\_about\_focus} &446 &    photography\_learn\_photographer\_art \\
533 &              \textbf{framing\_frame\_framed\_filled} &162 &                \textbf{focus\_subject\_main\_sharp} &440 &           \textbf{bw\_conversion\_color\_version} \\
526 &                    road\_where\_city\_place &154 &                 boat\_boats\_pier\_horizon &411 &              leaf\_leaves\_plant\_plants \\
484 & congratulations\_congrats\_proud\_fantastic &154 &                   looks\_good\_great\_very &409 &      \textbf{vignette\_vignetting\_heavy\_strong} \\
473 &           \textbf{tones\_tone\_tonemapping\_mapping} &137 &    building\_buildings\_right\_perspective &407 &       critique\_criticism\_critiques\_no \\
464 &    building\_buildings\_tower\_architecture &137 &               \textbf{horizon\_line\_frame\_middle} &406 &          rock\_rocks\_foreground\_bottom \\
462 &           \textbf{border\_borders\_fan\_distracting} &134 &                       dog\_dogs\_fur\_eyes &400 &         beautiful\_pic\_gorgeous\_lovely \\
431 &                  \textbf{focus\_out\_focused\_seems} &125 &        butterfly\_wings\_butterflies\_wing &390 & reflection\_mirror\_reflections\_mirrors \\
422 &                meets\_challenge\_meet\_fits &118 &          animal\_animals\_wildlife\_monkey &388 &    stars\_star\_trails\_astrophotography \\
417 &      \textbf{lighting\_light\_brighter\_composition} &112 &           \textbf{bw\_contrast\_choice\_conversion} &371 &     portrait\_portraits\_landscape\_self \\
\bottomrule
\end{tabular}}
\end{table}

\subsection{Informativeness Analysis}
\label{appx:info_score}

 We use the definition of informativeness score of a previous work~\cite{ghosal2019aesthetic} as a proxy of how meaningful are the comments in our dataset and how do they compare to other datasets. This definitions leverages on the relative frequency of unigrams and bigrams respect to the total vocabulary of the corpus. Then, a comment is represented as the union of its unigrams and bigrams and it is assigned an informativeness score $\rho$ as the average of the negative log probabilities of its unigrams ($P(u_{i})$) and bigrams ($P(b_{j})$):

\begin{equation}
\rho = -\frac{1}{2}[log\prod_{i}^{N}P(u_{i}) + log\prod_{j}^{M}P(b_{j})]
\end{equation}

\begin{figure}
    \centering
    \includegraphics[width=.8\linewidth]{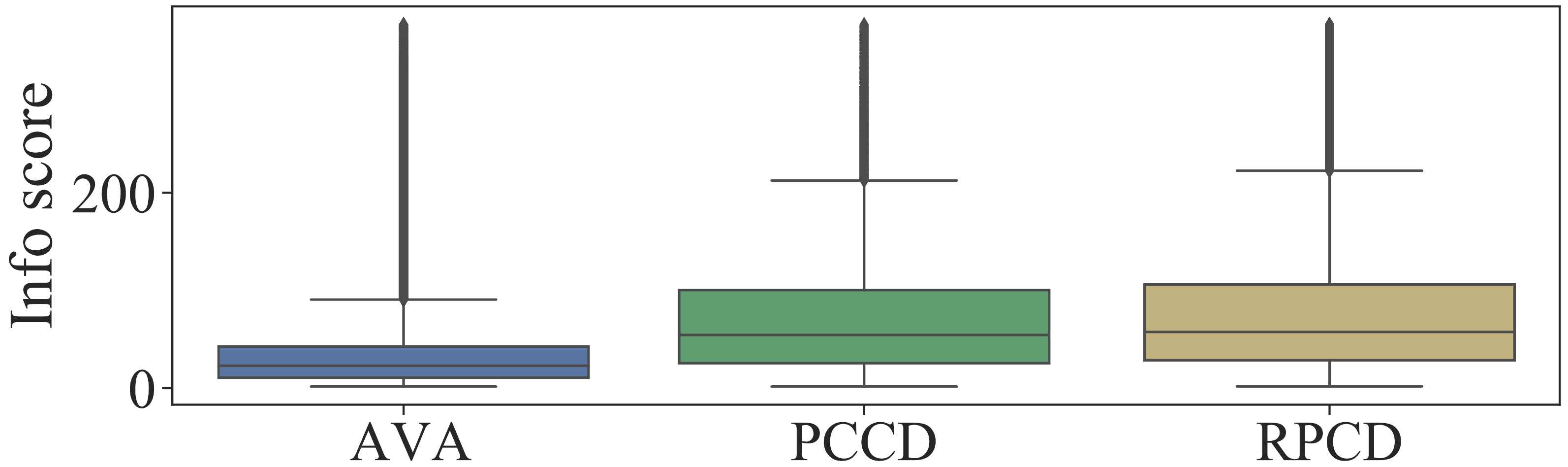}
    \caption{Informativeness score for each of the considered datasets.}
    \label{fig:info-score-comparison}
\end{figure}

The more frequent a word is, the less informative it will be. We observe that the proposed RPCD dataset have a slightly higher informativeness score than PCCD dataset, while both of them have a score twice as high as AVA dataset. The box plot shown in Figure~\ref{fig:info-score-comparison} describes the informativeness score distribution among the different datasets, where the dataset with the highest average informativeness score is RPCD (78.61), followed by PCCD (72.96) and finally, with less than half the score, AVA (32.43). 
%


\section{Experiments and Implementation Details}
\label{sec:implementation}

\subsection{Image Aesthetic Assessment}
\paragraph{AestheticViT.} For ranking the images with respect to aesthetic or sentiment scores, we experiment with different models based on Vision Transformer (ViT) \cite{dosovitskiy2020image} as a baseline as this architecture has proved its effectiveness for several tasks. We run experiments with various versions of ViT in terms of model size (i.e., Tiny, Small, Base, and Large), input patch size, and pre-training dataset. In what follows we use brief notation to indicate the model size and the input patch size: for instance, ViT-L/16 means the ``Large'' variant with $16\times16$ input patch size. We also consider the Data-efficient image Transformer (DeiT), a post-ViT model that improves the training process and performance.

On top of the pre-trained transformer, we add a fully-connected layer which is randomly initialized. The whole model is then trained to predict the final score using the mean square error as the loss function. We resize the input images to have the maximum input size of 700 pixels and adjust the other size to preserve the original aspect ratio. To handle images with varying resolutions, we scale input positional embedding accordingly with the image resolution by performing bilinear interpolation. During training we adopt a batch size of 1 because the images can have different resolutions. We finetune the models until convergence, for a maximum of 5 epochs (although convergence usually occurs on epoch 2 or 3). The learning rate is empirically set to 1$e$-6 and we use Adam optimizer. 
We exploit the available model implementations and pre-trained weights of the PyTorch Image Models library\footnote{\url{https://github.com/rwightman/pytorch-image-models}}.

We first perform experiments for estimating the aesthetic score of AVA \cite{murray2012ava} and PCCD \cite{jin2019aesthetic} (i.e., the only two datasets with both comments and aesthetic scores). Table \ref{tab:vit-aesthetic-score} reports performance on the test sets in terms of SRCC, and LCC for aesthetic score regression and accuracy for low-/high- aesthetic categorization. The accuracy is computed defining as high quality images those with an score above 5, and poor quality otherwise. The best results are achieved by the \texttt{ViT-L/16} pre-trained on ImageNet-21k and other considerations can be made. First, the use of larger patches, that is $32\times32$ pixels instead of $16\times16$ pixels, causes a significant drop in performance for the same model size. In fact, we have that \texttt{ViT-B/32} achieves 0.446 of SRCC, while \texttt{ViT-B/16} obtains 0.759 of SRCC on AVA. Second, the performance increases as the model size grows. On AVA, the SRCC is equal to 0.725 for \texttt{ViT-T/16} and 0.793 for \texttt{ViT-L/16}. Third, the DeiT models are slightly less performing than the basic ViT versions and pre-training on ImageNet-21k instead of ImageNet results in a minimal increase in results, i.e., about 0.02.


%
\begin{table}
    \centering
    \caption{Comparison of various transformers for image aesthetic assessment on AVA and PCCD. In each column, the best and second-best results are marked in \textbf{boldface} and \underline{underlined}, respectively.}
    \label{tab:vit-aesthetic-score}
    \begin{tabular}{lccccccc}
    \toprule
        \multirow{2}{*}{Model} & \multirow{2}{*}{Pretrain dataset} & \multicolumn{3}{c}{AVA} & \multicolumn{3}{c}{PCCD} \\
         &  & SRCC & LCC & Acc. (\%) & SRCC & LCC & Acc. (\%) \\ \midrule
        DeiT-T/16 & ImageNet & 0.725 & 0.731 & 80.33 & 0.227 & 0.262 & \textbf{98.34} \\
        DeiT-S/16 & ImageNet & 0.746 & 0.750 & 80.90 & 0.289 & 0.296 & \textbf{98.34} \\
        DeiT-B/16 & ImageNet & 0.765 & 0.768 & \underline{81.95} & 0.203 & 0.205 & \underline{98.22} \\
        ViT-S/16 & ImageNet & 0.734 & 0.738 & 81.00 & 0.277 & 0.293 & \underline{98.22} \\
        ViT-B/16 & ImageNet & 0.759 & 0.762 & 81.38 & \underline{0.297} & 0.318 & \textbf{98.34} \\
        ViT-B/32 & ImageNet & 0.446 & 0.464 & 73.93 & 0.059 & 0.075 & \textbf{98.34} \\
        ViT-B/16 & ImageNet-21k & \underline{0.773} & \underline{0.774} & 81.91 & 0.282 & \underline{0.322} & \textbf{98.34} \\
        ViT-L/16 & ImageNet-21k & \textbf{0.793} & \textbf{0.793} & \textbf{82.85} & \textbf{0.369} & \textbf{0.367} & \textbf{98.34} \\ \bottomrule
    \end{tabular}
\end{table}

Table \ref{tab:ava-aesthetic-comparison} reports the comparison with state-of-the-art methods on the AVA dataset (for PCCD, there is no benchmark for aesthetic score assessment). Our \texttt{ViT-L/16} pre-trained on ImageNet-21k (in the table named as \texttt{ViT-L/16 - \textit{21k}}) obtains better performance than Hosu \etal~\cite{hosu2019effective} for aesthetic score regression with an increment of 0.04 on both SRCC and LCC. On the other hand, we are in second place for the aesthetic classification with an accuracy of 0.35\% lower than Chen \etal~\cite{chen2020adaptive}. Correlation metrics are more adequate than accuracy \cite{hosu2019effective}, and exact score estimation is more challenging and representative of the full range of scores. Therefore, we can claim that we have achieved an excellent result.
%
\begin{table}
    \centering 
    \caption{Comparison of our baseline with state-of-the-art methods on the AVA dataset for image aesthetic assessment. In each column, the best and second-best results are marked in \textbf{boldface} and \underline{underlined}, respectively. The ``--'' means that the result is not available.}
    \label{tab:ava-aesthetic-comparison}
    \begin{tabular}{lccc}
    \toprule
        Model & SRCC & LCC & Accuracy (\%) \\ \midrule
        Murray \etal \cite{murray2012ava} & -- & -- & 66.70 \\
        Lu \etal \cite{lu2014rapid} & -- & -- & 74.46 \\
        Ma \etal \cite{ma2017lamp} & -- & -- & 81.70 \\
        Kong \etal \cite{kong2016photo} & 0.558 & -- & 77.33 \\
        Talebi \etal \cite{talebi2018nima} & 0.612 & 0.636 & 81.51 \\
        Chen \etal \cite{chen2020adaptive} &  0.649 & 0.671 & \textbf{83.20} \\
        Xu \etal \cite{xu2020deep} &  0.724 &  0.725 & 80.90 \\
        Ke \etal \cite{ke2021musiq} &  0.726 &  0.738 & 81.15 \\
        Celona \etal \cite{celona2021composition} &  0.731 &  0.732 & 80.75 \\
        Hosu \etal \cite{hosu2019effective} & \underline{0.756} & \underline{0.757}  & 81.72 \\
        ViT-L/16 - \textit{21k} & \textbf{0.793} & \textbf{0.793} & \underline{82.85} \\
        \bottomrule
    \end{tabular}
\end{table}

We perform experiments considering the previous backbones, apart from \texttt{ViT-B/32} which produced the worst results, for the sentiment score estimation. Results on AVA, PCCD and our RPCD are reported in Table \ref{tab:vit-sentiment-score}. We observe the same behavior as the aesthetic assessment, that the larger models outweigh the smaller ones. We also point out that \texttt{ViT-L/16 - \textit{21k}} achieves slightly higher performance than \texttt{ViT-L/16} on AVA, vice versa on RPCD. Finally, on PCCD we get the worst results in terms of correlation and the best results for classification compared to the other two datasets.
\begin{table}
    \centering
    \caption{Results obtained using ViT for estimating the sentiment score on AVA, PCCD, and RPCD. In each column, the best and second-best results are marked in \textbf{boldface} and \underline{underlined}, respectively.}
    \label{tab:vit-sentiment-score}
    \resizebox{\linewidth}{!}{\begin{tabular}{lccccccccc}
    \toprule
    \multirow{2}{*}{Model} & \multicolumn{3}{c}{AVA} & \multicolumn{3}{c}{PCCD} & \multicolumn{3}{c}{RPCD} \\
     & SRCC & LCC & Acc. (\%)  & SRCC & LCC & Acc. (\%) & SRCC & LCC & Acc. (\%) \\ \midrule
    DeiT-T/16 & 0.492 & 0.507 & 90.46 & 0.187 & 0.220 & \textbf{93.87} & 0.188 & 0.189 & 64.68 \\
    DeiT-S/16 & 0.500 & 0.513 & 90.48 & 0.170 & 0.182 & \textbf{93.87} & 0.190 & 0.189 & 64.61 \\
    DeiT-B/16 & 0.529 & 0.535 & \underline{90.53} & 0.202 & 0.233  &     \textbf{93.87} & 0.216 & 0.218 & 64.62 \\ 
    ViT-S/16  & 0.498 & 0.512 & 90.41 & 0.192 & 0.211 & \underline{93.75} & 0.202 & 0.199 & 64.65 \\
    ViT-B/16 & 0.527 & 0.534 & 90.46 & \textbf{0.228} & \textbf{0.262} & \textbf{93.87} & 0.230 & 0.230 & 65.00 \\
    ViT-L/16 & \underline{0.542} & \textbf{0.551} & 90.50 & \underline{0.212} & \underline{0.236} & \textbf{93.87} & \textbf{0.249} & \textbf{0.253} & \textbf{65.27} \\ 
    ViT-B/16 - \textit{21k} & 0.533 & 0.534 & 90.47 & 0.206 & 0.225 & \textbf{93.87} & 0.228 & 0.228 & 64.73 \\
    ViT-L/16 - \textit{21k} & \textbf{0.544} & \underline{0.550} & \textbf{90.55} & 0.199 & 0.225 & \textbf{93.87} & \underline{0.246} & \underline{0.246} & \underline{65.08} \\
    \bottomrule
    \end{tabular}}
\end{table}
%

\paragraph{ViT + Linear probe.} The goal of this experiments is to assess to what extent the results obtained to predict the aesthetic and sentiment scores are due to the knowledge already present in the pre-trained model. We use the pre-trained ViT models as feature extractors and then we fit a linear regressor on those extracted features to predict the aesthetic score. This linear regressor was implemented as a Stochastic Gradient Descent Regressor with Scikit-Learn~\cite{scikit-learn}. In Table \ref{tab:linear-probe-aesthetic-score} are reported the results for image aesthetic assessment on AVA and PCCD.
\begin{table}
    \centering
    \caption{Results obtained by using ViT as a feature extractor followed by a linear regressor (we called ViT + Linear probe) for estimating the aesthetic score on AVA and PCCD. In each column, the best and second-best results are marked in \textbf{boldface} and \underline{underlined}, respectively.}
    \label{tab:linear-probe-aesthetic-score}
    \begin{tabular}{lcccccc}
    \toprule
        \multirow{2}{*}{Model} & \multicolumn{3}{c}{AVA} & \multicolumn{3}{c}{PCCD} \\
         & SRCC & LCC & Acc. (\%) & SRCC & LCC & Acc. (\%) \\ \midrule
        DeiT-T/16 & 0.345 & 0.355 & 71.66 & 0.185 & 0.191 & \underline{98.34} \\
        DeiT-S/16 & 0.454 & 0.459 & 74.27 & 0.212 & 0.203 & \underline{98.34} \\
        DeiT-B/16 & 0.506 & 0.510 & 74.89 & 0.203 & 0.205 & 98.22\\
        ViT-S/16 & 0.484 & 0.489 & 74.60 & 0.163 & 0.189 & \underline{98.34}\\
        ViT-B/16 & \underline{0.553} & \underline{0.557} & \underline{75.69} & \textbf{0.254} & \textbf{0.272} & 97.98 \\
        ViT-L/16 & 0.528 & 0.534 & 74.73 & 0.203 & 0.222 & \textbf{98.46} \\
        ViT-B/16 - \textit{21k} & \textbf{0.570} & \textbf{0.570} & \textbf{76.44} & \underline{0.241} & \underline{0.246} & \underline{98.34} \\
        ViT-L/16\ - \textit{21k} & 0.502 & 0.505 & 74.48 & 0.210 & 0.222 & \textbf{98.46} \\
        \bottomrule
        \end{tabular}
\end{table}
Table \ref{tab:linear-probe-sentiment-score} presents the results of the same experiment but using the sentiment score instead. Table~\ref{tab:aesth_score_difference} and Table~\ref{tab:sent_score_difference} show the difference in performance between the trained models and the linear probe experiments. We can observe how for every case and every metric (except for the accuracy of Vit-S and ViT-L-21k on PCCD dataset to predict the aesthetic score), training the models outperform the pre-trained models (linear probes). The increase in performance is higher on AVA dataset, while PCCD and RPCD datasets do not benefit that much of further training. This may suggest that there is room for better training procedures on this datasets.
\begin{table}
    \centering
    \caption{Results obtained by using ViT as a feature extractor followed by a linear regressor (we called ViT + Linear probe) for estimating the sentiment score on the three considered datasets. In each column, the best and second-best results are marked in \textbf{boldface} and \underline{underlined}, respectively.}
    \label{tab:linear-probe-sentiment-score}
    \resizebox{\linewidth}{!}{\begin{tabular}{lccccccccc}
    \toprule
    \multirow{2}{*}{Model} & \multicolumn{3}{c}{AVA} & \multicolumn{3}{c}{PCCD} & \multicolumn{3}{c}{RPCD} \\
     & SRCC & LCC & Acc. (\%)  & SRCC & LCC & Acc. (\%) & SRCC & LCC & Acc. (\%) \\ \midrule
    DeiT-T/16 & 0.238 & 0.235 & 90.26 & 0.153 & 0.151 & \textbf{93.87} & 0.107 & 0.108 & 62.56 \\
    DeiT-S/16 & 0.300 & 0.303 & 90.27 & 0.139 & 0.135 & \underline{93.40} & 0.128 & 0.128 & 63.09 \\
    DeiT-B/16 & 0.338 & 0.342 & \textbf{90.32} & 0.136 & 0.127 & 93.16 & 0.129 & 0.129 & 63.74 \\
    ViT-S/16 & 0.320 & 0.322 & \underline{90.30} & \underline{0.152} & \underline{0.162} & 92.22 & 0.115 & 0.115 & 61.88 \\
    ViT-B/16 & \underline{0.369} & \underline{0.375} & 90.27 & 0.131 & \textbf{0.166} & 93.04 & 0.144 & 0.142 & 61.02 \\
    ViT-L/16 & 0.366 & 0.366 & 90.26 & \textbf{0.156} & \textbf{0.166} & 93.04 & 0.136 & 0.140 & 62.48 \\ 
    ViT-B/16 - \textit{21k} & \textbf{0.392} & \textbf{0.395} & 90.27 & 0.111 & 0.114 & \underline{93.40} & \textbf{0.172} & \textbf{0.174} & \textbf{64.59} \\
    ViT-L/16 - \textit{21k} & 0.348 & 0.348 & 90.26 & 0.145 & 0.158 & \underline{93.40} & \underline{0.154} & \underline{0.155} & \underline{64.44} \\
    \bottomrule
    \end{tabular}}
\end{table}
%

\begin{table}
\centering
\caption{Performance difference between ViT + Linear Probe and Aesthetic ViT (Table~\ref{tab:vit-aesthetic-score} - Table~\ref{tab:linear-probe-aesthetic-score}) for aesthetic score.}
\label{tab:aesth_score_difference}
\begin{tabular}{lcccccc}
\hline

\multirow{2}{*}{Model}                               & \multicolumn{3}{c}{AVA}     & \multicolumn{3}{c}{PCCD}    \\
                                                     & SRCC   & LCC    & Acc. (\%) & SRCC   & LCC    & Acc. (\%) \\ \hline

DeiT-T/16                                            & +0.380 & +0.376 & +8.670    & +0.042 & +0.071 & +0.000    \\
DeiT-S/16                                            & +0.292 & +0.291 & +6.630    & +0.077 & +0.093 & +0.000    \\
DeiT-B/16                                            & +0.259 & +0.258 & +7.060    & +0.000 & +0.000 & +0.000    \\
ViT-S/16                                             & +0.250 & +0.249 & +6.400    & +0.114 & +0.104 & -0.120    \\
ViT-B/16                                             & +0.206 & +0.205 & +5.690    & +0.043 & +0.046 & +0.360    \\
ViT-B/16 - \textit{21k}               & +0.203 & +0.204 & +5.470    & +0.041 & +0.076 & +0.000    \\
ViT-L/16 - \textit{21k} & +0.291 & +0.288 & +8.370    & +0.159 & +0.145 & -0.120   \\
\bottomrule

\end{tabular}
\end{table}

\begin{table}
\centering
\caption{Performance difference between ViT + Linear Probe   and Aesthetic ViT (Table~\ref{tab:vit-sentiment-score} - Table~\ref{tab:linear-probe-sentiment-score}) for sentiment score.}
\label{tab:sent_score_difference}
\resizebox{\linewidth}{!}{\begin{tabular}{lccccccccc}
\hline
\multirow{2}{*}{Model}               & \multicolumn{3}{c}{AVA}     & \multicolumn{3}{c}{PCCD}    & \multicolumn{3}{c}{RPCD}    \\
                                     & SRCC   & LCC    & Acc. (\%) & SRCC   & LCC    & Acc. (\%) & SRCC   & LCC    & Acc. (\%) \\ \hline
DeiT-T/16                            & +0.254 & +0.272 & +0.200    & +0.034 & +0.069 & +0.000    & +0.081 & +0.081 & +2.120    \\
DeiT-S/16                            & +0.200 & +0.210 & +0.210    & +0.031 & +0.047 & +0.470    & +0.062 & +0.061 & +1.520    \\
DeiT-B/16                            & +0.191 & +0.193 & +0.210    & +0.066 & +0.106 & +0.710    & +0.087 & +0.089 & +0.880    \\
ViT-S/16                             & +0.178 & +0.190 & +0.110    & +0.040 & +0.049 & +1.530    & +0.087 & +0.084 & +2.770    \\
ViT-B/16                             & +0.158 & +0.159 & +0.190    & +0.097 & +0.096 & +0.830    & +0.086 & +0.088 & +3.980    \\
ViT-L/16                             & +0.176 & +0.185 & +0.240    & +0.056 & +0.070 & +0.830    & +0.113 & +0.113 & +2.790    \\
ViT-B/16 - \textit{21k} & +0.141 & +0.139 & +0.200    & +0.095 & +0.111 & +0.470    & +0.056 & +0.054 & +0.140    \\
ViT-L/16 - \textit{21k} & +0.196 & +0.202 & +0.290    & +0.054 & +0.067 & +0.470    & +0.092 & +0.091 & +0.640    \\ \hline
\end{tabular}}
\end{table}

\paragraph{NIMA.} We compare the previous ViT models with a model from the literature, i.e., NIMA~\cite{talebi2018nima}, for sentiment score prediction. NIMA is trained by us using the code released by its authors. We use an ImageNet-trained VGG-16 as the backbone. Input images are resized to a fixed spatial resolution of $224 \times 224$ pixels. As in \cite{talebi2018nima}, for model optimization we exploit the Earth Mover's Distance (EMD):

\begin{equation}
  EMD(\mathsf{\hat{q}},\mathsf{q}) = \left(\frac{1}{N} \sum_{k=1}^{N}{|CDF_{\mathsf{\hat{q}}}(k)-CDF_{\mathsf{q}}(k)|^r} \right)^{\frac{1}{r}},
\label{eq:emd}
\end{equation}
where $\mathsf{\hat{q}}$ and $\mathsf{q}$ are the ground-truth and the predicted score distributions, respectively. Finally, $CDF_*(k)$ is the cumulative distribution function, $r$ equal to 2 is used to penalize the Euclidean distance between the CDFs. We use the probability distribution on the three sentiment polarity classes $\mathsf {p}$ as ground-truth. We optimize the model by using Stochastic Gradient Descent (SGD) with learning rate of 5$e$-3 and batch size equal to 64 for 100 epochs. We use an early stopping policy based on validation loss with a patience term of 10 epochs.

\paragraph{Summary.}
Experiments with \texttt{ViT + Linear probe} have shown that pre-trained ViTs for image recognition do not work well for predicting aesthetic and sentiment scores. It is therefore necessary to train the model to learn the characteristics that best encode the various aspects of aesthetics. Table \ref{tab:aesth_score_difference} and Table \ref{tab:sent_score_difference} report the difference in performance between \texttt{ViT + Linear prob} and \texttt{AestheticViT} models for aesthetic score estimation and sentiment score estimation, respectively. This way we highlight the gain obtained thanks to the training of the backbones.

Among the various tested models, \texttt{ViT-L/16 - \textit{21k}} achieved the best results on both AVA and PCCD for aesthetic assessment. It also outperforms state-of-the-art aesthetic assessment methods on the AVA dataset. On the other hand, for the prediction of the sentiment score the \texttt{ViT-L/16} model obtained the best performance regardless of the dataset used for pre-training.
\subsection{Image Aesthetic Critique Generation}
\label{appx:aesth_critiques}
We verify the use of the proposed dataset for the generation of aesthetic image critique by using Bootstrapping Language Image Pre-training (BLIP) \cite{li2022blip}. It is a method for the unified understanding and generation of the visual language. A pre-trained ViT-B/16 on the COCO dataset is finetuned for aesthetic captioning by exploiting the AdamW optimizer with initial learning rate equal to 1$e$-5, weight decay of 0.05, and a cosine learning rate schedule. We train for 5 epochs using a batch size of 16 samples. During inference, we use beam search with a beam size of 3, and set the minimum and maximum generation lengths as 20 and 50, respectively.

\section{Resources Used}
\label{sec:resources}

In this section we briefly list the resources used to carry out this work:
\begin{itemize}
    \item Host machines: The machines used by the authors, each of them with access to a GPU NVIDIA GeForce RTX 2080 Ti.
    \item Access to internal cluster\footnote{\url{https://scicomp.ethz.ch/wiki/Euler}} with access to various instances with the following GPUs: NVIDIA GeForce RTX 2080 Ti and NVIDIA TITAN RTX.
    \item A part of the experiments, but not all, were logged to Weights \& Biases\footnote{\url{https://wandb.ai/}}, which registered the time used for those experiments, summing up a total of ~2500 hours.
\end{itemize}

\section{Ethical considerations}
\label{sec:ethics}

This section comments on the Ethics Guidelines~\footnote{\url{https://nips.cc/public/EthicsGuidelines}} of NeuroIPS. In particular, we comment on various of the points brought on this guidelines:
\begin{itemize}
    \item \textbf{Personally identifiable information and data collection}. The samples in our dataset are attached to the user ID. While this provides a first level of anonymity to the users, it is fairly straight forward to access the public user profile, which may contain identifiable information the user had previously agreed to share and may be identifiable. Every user consents the collection of this information and accepts the Reddit's Privacy Policy~\footnote{\url{https://www.reddit.com/policies/privacy-policy}}, where it is stated that \textit{"[...] Reddit also allows third parties to access public Reddit content via the Reddit API and other similar technologies. [...]"}. Thus, not every user has been directly asked for consent to include data produced by them in this dataset, but this consent is comprised under the Privacy Policy and the Reddit API terms of Use. We expand on this in the Section~\ref{app:datasheet}. However, we point out that, a priori, disclosing that a person has any activity or belongs to the \texttt{r/photocritique} subreddit does not involve degrading or embarrassing such person.
    
    \ADD{\item \textbf{Data consent}. As pointed out above, every user consents accepts the Reddit's Privacy Policy, where it is stated that \textit{"[...] Reddit also allows third parties to access public Reddit content via the Reddit API and other similar technologies. [...]"}. The use of Reddit as a source of data for a large variety of scientific research has had an important impact in several fields as described in The Pushshift Dataset work~\cite{baumgartner2020pushshift}. We acknowledge that there is not explicit consent of the users to use their data for scientific purposes. However, we considered this to be covered by Reddit's Privacy Policy. Hence, instead of collecting and storing the metadata and data produced by users, we provide the identifiers necessary to access the data and the tools to construct the dataset.}
    
    \item  \textbf{Explicit content}. Images may contain explicit content of people. The first of the community rules state \textit{1. Post only photos you took. Do not post a photo unless you took it! [...]} Thus, it is assumed this rule implies that the user posting a new image is the owner of the photography and hence has the right to distribute it. The sensitive content is labeled as "NSFW" in the dataset.
    
    \item \textbf{Bias against people of a certain gender, race, sexuality, or who have other protected characteristics}. This is a multi-factor issue that must be addressed from different perspectives and is beyond the scope of the first analyses presented in this paper to show the usability of this new data source. For instance, questions such as the impact of gender, race or sexuality on the perceived aesthetics of an image or how these images are critiqued are completely out of the scope of this work. However, we must note and acknowledge the work of the team of moderators of the \texttt{r/photocritique} community. Not only they approve each of the posts published in the community, but it is clearly stated that inappropriate or disrespectful posts are banned. As stated in the rules of the community: \textit{Lewd comments or those deemed by the moderation team to be grossly inappropriate will result in a permanent ban. You have been warned.}. And as stated in the critiques guidelines: \textit{We do not allow [...] inappropriate/sexist/racist comments.}.
    
    \item \textbf{Filtering of offensive content}. Due to the scale of the dataset, it has not been feasible to double check every post complies with the community rules. \ADD{However, we have included a preliminary analysis of the presence of offensive content in the dataset (See Appendix~\ref{appx:offensive}), in which we found that the predicted offensive content in the comments of the dataset is under 4\%.}

\end{itemize}

\section{License}
\label{sec:License}

We comply with Reddit User Agreement\footnote{\url{https://www.redditinc.com/policies/user-agreement/}}, Reddit API terms of use~\footnote{\url{https://docs.google.com/a/reddit.com/forms/d/e/1FAIpQLSezNdDNK1-P8mspSbmtC2r86Ee9ZRbC66u929cG2GX0T9UMyw/viewform}} and PushShift database Creative Commons License~\footnote{\url{https://zenodo.org/record/3608135\#.Yp3XEXZBw2w}}. In particular, we refer to the Section 2.d of Reddit API Terms of Use, which states: "User Content.  Reddit user photos, text and videos ("User Content") are owned by the users and not by Reddit. Subject to the terms and conditions of these Terms, Reddit grants You a non-exclusive, non-transferable, non-sublicensable, and revocable license to copy and display the User Content using the Reddit API through your application, website, or service to end users.  You may not modify the User Content except to format it for such display. You will comply with any requirements or restrictions imposed on usage of User Content by their respective owners, which may include "all rights reserved" notices, Creative Commons licenses or other terms and conditions that may be agreed upon between you and the owners." We do not provide access to any data directly, but a list of IDs associated with a post on Reddit. This information is then used to retrieve the images, comments and metadata using the provided tools after obtaining a license key for the official Reddit API. Moreover, we do not modify the original content by no means, while we provide the necessary tools to process the data and run the same experiments we carried out. 

We release the dataset under the Creative Commons Attribution 4.0 International license.

\ADD{\section{Datasheet for RPCD}
\label{app:datasheet}
In this section we detail the datasheet presented in \cite{gebru2021datasheets} for documenting the proposed dataset. Note that, while we do not provide any data other that the IDs associated to Reddit posts, we answer the questionnaire considering the constructed dataset resulted from using our code.

\subsection{Motivation}

\begin{itemize}
    \item \textbf{For what purpose was the dataset created?} 
    
    RPCD was created to drive the research progress in both image aesthetic assessment and aesthetic image captioning. The proposed dataset addresses the need for images acquired with modern acquisition devices and photo critiques that give a better understanding of how the aesthetic evaluation is carried out.
    
    \item \textbf{Who created this dataset (e.g., which team, research group) and on behalf of which entity (e.g., company, institution, organization)?}
    
    This dataset was created by the authors on behalf of their respective institutions, ETH Media Technology Center and University of Milano-Biococca.
    
    \item \textbf{Who funded the creation of the dataset?} 
    
    The creation of this dataset was carried out as part of the Aesthetic Assessment of Image and Video Content project\footnote{\url{https://mtc.ethz.ch/research/image-video-processing/aesthetics-assessment.html}}. The project is supported by Ringier, TX Group, NZZ, SRG, VSM, viscom, and the ETH Zurich Foundation on the ETH MTC side.
\end{itemize}

\subsection{Composition}

\begin{itemize}
    \item \textbf{What do the instances that comprise the dataset represent (e.g., documents, photos, people, countries)?} 
    
    Each instance is represented as a tuple containing one image and several photo critiques, where the images are JPEG files and the photo critiques are in textual form.

    \item \textbf{How many instances are there in total (of each type, if appropriate)?}
    
    RPCD consists of 73,965 data instances. Specifically, there are 73,965 images and 219,790 photo critiques.

    \item \textbf{Does the dataset contain all possible instances or is it a sample (not necessarily random) of instances from a larger set?} 
    
    The dataset contains all samples (posts) available at the moment of collection, from the origin of the forum until the moment of collection. Additionally, included posts had to meet the following criteria:
    \begin{itemize}
        \item The post has at least an image which could be retrieved.
        \item The post has at least one comment critiquing the image
        \item The post is not a discussion thread, a type of post to encourage general discussion in the forum.
    \end{itemize}
    
    \item \textbf{What data does each instance consist of?} 
    
    Each data instance consists of an image and one or more textual photo critiques.
    
    \item \textbf{Is there a label or target associated with each instance? If so, please provide a description.} 
    
    There is no label associated with each sample. However, in this work we propose a method to compute said label, which is calculated using the processing scripts.
    
    \item \textbf{Is any information missing from individual instances?} 
    
    Some of the samples in the dataset might be missing at the moment of future retrievals due to the users removing the data from Reddit.
    
    \item \textbf{Are relationships between individual instances made explicit (e.g., users' movie ratings, social network links)?} 
    
    Every image and comment in the dataset is associated with the user who created the post. Moreover, we build the tree of comments of the different users criticizing an image. However, the data is downloaded by using only the post IDs.
    
    \item \textbf{Are there recommended data splits (e.g., training, development/validation, testing)?} 
    
    We provide the data splits we used in our experiments in the repository and they are used to retrieve the posts we used, although we encourage the use of other splits. The splits were randomly generated to divide the dataset in 70\% train, 10\% validation and 20\% test splits.
    
    \item \textbf{Are there any errors, sources of noise, or redundancies in the dataset?} 
    
    The source of data itself could be considered a source of noise. Additionally, we have not evaluated the case in which an image is posted by an user several times in different posts, although we consider this event to be non-existent or insignificant.
    
    \item \textbf{Is the dataset self-contained, or does it link to or otherwise rely on external resources (e.g., websites, tweets, other datasets)?} 
    
    The dataset links to resources available on Reddit and Pushshift. In particular, posts and their metadata (including the URLs to images) are retrieved from Pushshift, while the comments are retrieved directly from Reddit. There is no guarantee that the dataset will remain constant, as this depends on the users exercising their rights to remove their content from the dataset sources. For this same reason, there are not any archival versions of the complete dataset available online. In order to retrieve the dataset in the future, Reddit API credentials are needed. Please, refer to the instructions about how to obtain the credentials\footnote{\url{https://www.reddit.com/wiki/api/}}.
    
    \item \textbf{Does the dataset contain data that might be considered confidential (e.g., data that is protected by legal privilege or by doctor-patient confidentiality, data that includes the content of individuals' non-public communications)?} 
    
    The dataset does not contain any confidential data as both images and comments are publicly available in Reddit.
    
    \item \textbf{Does the dataset contain data that, if viewed directly, might be offensive, insulting, threatening, or might otherwise cause anxiety?} 
    
    There are data samples depicting explicit nudity with aesthetics purposes, and we acknowledge that this may be problematic for some people. According to the subreddit rules, this content must be marked: \textit{"Not Suitable for Work (NSFW) must be marked. [...] Please keep NSFW posts respectful. Nothing that would be considered pornography."} For this reason, the dataset processing script creates a NSFW column in the dataframe to easily filter this content. 
    
    \item \textbf{Does the dataset relate to people?} 
    
    Yes, some of the images contain people or the main subject is a person. 
    
    \item \textbf{Does the dataset identify any subpopulations (e.g., by age, gender)? } 
    
    No.
    
    \item \textbf{Is it possible to identify individuals (i.e., one or more natural persons), either directly or indirectly (i.e., in combination with other data) from the dataset?} 
    
    All posts and comments are linked to users, which may be identifiable depending on the data made available by the user. Additionally, posts and comments may contain information linking to other social media which could serve to identify a certain user.

    \item \textbf{Does the dataset contain data that might be considered sensitive in any way (e.g., data that reveals racial or ethnic origins, sexual orientations, religious beliefs, political opinions or union memberships, or locations; financial or health data; biometric or genetic data; forms of government identification, such as social security numbers; criminal history)?} 
    
    The retrieved data might contain sensitive data publicly disclosed by the users. However, we do not expect this to be common at all, and we would be surprised that some kinds of sensitive information are present in the community (financial, health, biometric, genetic or governmental data).
    
\end{itemize}

\subsection{Collection process}

\begin{itemize}
    \item \textbf{How was the data associated with each instance acquired?} 
    
    The data was directly observable (posts in Reddit stored in Pushshift's and Reddit's servers).
    
    \item \textbf{What mechanisms or procedures were used to collect the data (e.g., hardware apparatus or sensor, manual human curation, software program, software API)?} 
    
    Software API to access both Reddit and Pushshift. 
    
    \item \textbf{If the dataset is a sample from a larger set, what was the sampling strategy (e.g., deterministic, probabilistic with specific sampling probabilities)?} 
    NA.
    
    \item \textbf{Who was involved in the data collection process (e.g., students, crowdworkers, contractors) and how were they compensated (e.g., how much were crowdworkers paid)?} 
    
    Nobody was involved in the data collection process since all data was already available and observable.
    
    \item \textbf{Over what timeframe was the data collected?} 
    
    Tha data was collected in February 2022, and comprises posts and comments in the span from May 2009 (first posts in the subreddit) to February 2022 (collection date).
    
    \item \textbf{Were any ethical review processes conducted (e.g., by an institutional review board)?} 
    
    No ethical review process was conducted previous to the ethical review of this conference.
    
    \item \textbf{Does the dataset relate to people?} 
    
    Yes, but not exclusively. 
    
    \item \textbf{Did you collect the data from the individuals in question directly, or obtain it via third parties or other sources (e.g., websites)?} 
    
    Third party sources (Reddit and Pushshift).
    
    \item \textbf{Were the individuals in question notified about the data collection?} 
    
    No.
    
    \item \textbf{Did the individuals in question consent to the collection and use of their data?} 
    
    According to Reddit's Privacy Policy\footnote{\url{https://www.reddit.com/policies/privacy-policy}}, which is accepted by every user upon registration, \textit{"Reddit also allows third parties to access public Reddit content via the Reddit API and other similar technologies." }. Moreover, we note that no data from the users is made directly available in the dataset. It only contains the IDs of the posts and the tools to retrieve them from Reddit and Pushshift.
    
    \item \textbf{If consent was obtained, were the consenting individuals provided with a mechanism to revoke their consent in the future or for certain uses?} 
    
    Users may remove their data from Reddit and Pushshift using their respective privacy enforcing mechanisms. Thus, they would be removing their data from the dataset.
    
    \item \textbf{Has an analysis of the potential impact of the dataset and its use on data subjects (e.g., a data protection impact analysis) been conducted?} 
    
    As we note above, no data from the users is made directly available in the dataset. The dataset only contains the IDs of the posts and the tools to retrieve them from Reddit and Pushshift.
\end{itemize}

\subsection{Processing/cleaning/labeling}

\begin{itemize}
    \item \textbf{Was any preprocessing/cleaning/labeling of the data done (e.g., discretization or bucketing, tokenization, part-of-speech tagging, SIFT feature extraction, removal of instances, processing of missing values)?} 
    
    We provide scrips to automatically process the downloaded raw posts. Only first level comments are kept, posts with no comments or whose image is no longer available are filtered. 
    
    \item \textbf{Was the "raw" data saved in addition to the preprocessed/cleaned/labeled data (e.g., to support unanticipated future uses)?} 
    
    The raw posts need to be downloaded for further processing.
    
    \item \textbf{Is the software used to preprocess/clean/label the instances available?} 
    
    Yes. The software for downloading and preparing the dataset is available on our GitHub repository \footnote{\label{githubrepo}\url{https://github.com/mediatechnologycenter/aestheval}}.
    
\end{itemize}

\subsection{Uses}

\begin{itemize}
    \item \textbf{Has the dataset been used for any tasks already?}
    
    RPCD is introduced and used in the paper \texttt{Understanding Aesthetics with Language: A Photo Critique Dataset for Aesthetic Assessment}.
    
    
    \item \textbf{Is there a repository that links to any or all papers or systems that use the dataset?} 
    
    Papers using RPCD will be listed on the PapersWithCode web page\footnote{\url{https://paperswithcode.com/dataset/rpcd}}.
    
    \item \textbf{What (other) tasks could the dataset be used for?} 
    
    RPCD can be used for modelling works in the areas of knowledge retrieval and multimodal reasoning.
    
    \item \textbf{Is there anything about the composition of the dataset or the way it was collected and preprocessed/cleaned/labeled that might impact future uses?} 
    
    No, there are no known risks to the best of our knowledge.
    
    \item \textbf{Are there tasks for which the dataset should not be used?} 
    
    RPCD should not be used for automatically judging a photographer's skills based on the photo critiques. The latter, in fact, are to be understood as highly subjective judgments that depend on the emotions and background of the commentators and could go beyond the mere technical evaluation of the shot.
\end{itemize}

\subsection{Distribution}

\begin{itemize}
    \item \textbf{Will the dataset be distributed to third parties outside of the entity (e.g., company, institution, organization) on behalf of which the dataset was created?} 
    
    Yes, the dataset is made publicly accessible.
    
    \item \textbf{How will the dataset will be distributed (e.g., tarball on website, API, GitHub)?} 
    
    See our GitHub repository \footref{githubrepo} for downloading instructions. RPCD has the following DOI: \texttt{10.5281/zenodo.6985507}.
    
    \item \textbf{When will the dataset be distributed?}
    
    RPCD will be released to the public in August 2022.
    
    \item \textbf{Will the dataset be distributed under a copyright or other intellectual property (IP) license, and/or under applicable terms of use (ToU)?} 
    
    We release the dataset under the Creative Commons Attribution 4.0 International license\footnote{\url{https://creativecommons.org/licenses/by/4.0/}}.
    
    \item \textbf{Have any third parties imposed IP-based or other restrictions on the data associated with the instances?} 
    
    No.
    
    \item \textbf{Do any export controls or other regulatory restrictions apply to the dataset or to individual instances?} 
    
    No.
    
\end{itemize}

\subsection{Maintenance}

\begin{itemize}
    \item \textbf{Who is supporting/hosting/maintaining the dataset?}
    
    RPCD is supported and maintained by ETH MTC and University of Milano-Bicocca. The post IDs are available on Zenodo, the posts are on Reddit and Pushshift, and the code for automatically retrieving the posts is on GitHub.
    
    \item \textbf{How can the owner/curator/manager of the dataset be contacted (e.g., email address)?}
    
    By emailing to \{daniel.veranieto,clabrador\}@inf.ethz.ch or luigi.celona@unimib.it. By opening an issue on our GitHub repository~\footref{githubrepo}.
    
    \item \textbf{Is there an erratum?} 
    
    All changes to the dataset will be announced on our Zenodo repository \footnote{\url{https://doi.org/10.5281/zenodo.6985507}\label{zenodorepo}}.

    \item \textbf{Will the dataset be updated (e.g., to correct labeling errors, add new instances, delete instances')?} 
    
    All updates (if necessary) will be posted on our Zenodo repository \footref{zenodorepo}.
    
    \item \textbf{If the dataset relates to people, are there applicable limits on the retention of the data associated with the instances (e.g., were individuals in question told that their data would be retained for a fixed period of time and then deleted)? } 
    
    The data related to users is stored on Reddit and Pushshift servers, and their data retention policies apply.
    
    \item \textbf{Will older versions of the dataset continue to be supported/hosted/maintained?} 
    
    All changes to the dataset will be announced on our Zenodo repository \footref{zenodorepo}. Outdated versions will be kept around for consistency.
    
    \item \textbf{If others want to extend/augment/build on/contribute to the dataset, is there a mechanism for them to do so?} 

    Any extension/augmentation by an external party is allowed under the release license. The dataset could be easily extended with other communities and other time periods using the available scripts. In order to add the extended version to the existing repositories, please contact the authors.
\end{itemize}
}

\end{document}